\documentclass{article}

\ifdefined\pdfsuppressptexinfo
\fi

\usepackage[preprint]{neurips_2026}

\usepackage[utf8]{inputenc} % allow utf-8 input
\usepackage[T1]{fontenc}    % use 8-bit T1 fonts
\usepackage{hyperref}       % hyperlinks
\usepackage{url}            % simple URL typesetting
\usepackage{booktabs}       % professional-quality tables
\usepackage{amsfonts}       % blackboard math symbols
\usepackage{nicefrac}       % compact symbols for 1/2, etc.
\usepackage{microtype}      % microtypography
\usepackage{xcolor}         % colors
\usepackage{amsmath}
\usepackage{graphicx}
\usepackage{subcaption}

\title{Still: Amortized KV Cache Compaction in a Single Forward Pass}

\author{%
  Charles O'Neill\thanks{First author.} \\
  Baseten \\
  San Francisco \\
  \texttt{charlie.oneill@baseten.co} \\
  \And
  Alex Sandomirsky\footnotemark[1] \\
  Baseten \\
  San Francisco \\
  \texttt{alex.sandomirsky@baseten.co} \\
  \And
  Harry Partridge\footnotemark[1] \\
  Baseten \\
  San Francisco \\
  \texttt{harry.partridge@baseten.co} \\
  \And
  Mudith Jayasekara \\
  Baseten \\
  San Francisco \\
  \texttt{mudith@baseten.co} \\
  \And
  Max Kirkby \\
  Baseten \\
  London \\
  \texttt{max.kirkby@baseten.co} \\
}

\begin{document}

\maketitle

\begin{abstract}
The KV cache is the memory bottleneck of long-horizon language model deployment. Practically, a deployable compactor must be lightweight enough to call during inference, expressive enough to preserve context under constraint, and reusable across a trajectory. Existing compaction methods satisfy only part of this requirement: selection methods are lightweight but subset-bound, while synthesis methods are expressive but rely on per-context optimization. Here we introduce Still, a small per-layer Perceiver trained once against a frozen base model that produces compact keys and values in a single forward pass. On Qwen and Gemma models, Still occupies the favorable side of the speed--quality frontier across compression ratios from $8\times$ to $200\times$ and context lengths from $8$k to $128$k. On the long-context RULER grid, Still exceeds the strongest baseline by 8--22 points. The same compact cache also supports free-form summarization, preserving most of the full-context gain on HELMET and winning a pairwise LongBench summarization comparison against KV-Distill. Because compaction is a forward pass, Still can be applied iteratively, entering a long-horizon regime unavailable to per-context methods. We show that amortization makes long-context cache compaction tractable, and synthesis makes its compact state useful at extreme compression.
\end{abstract}

\section{Introduction}

The KV cache is the memory bottleneck of long-horizon language model
deployment \citep{kwon2023efficient}. As LLMs operate over multi-day
coding agents, multi-turn tool use, and repository-scale reasoning,
the cache becomes the binding constraint on what these systems can
do. Today's options are an all-or-nothing choice: a lossless cache
that grows linearly with context length, or lossy alternatives
(fine-tuning, retrieval-augmented generation, document summaries)
that abandon the structure and fidelity of the model's own internal
representations.
 
\begin{figure}
    \centering
    \includegraphics[width=0.85\linewidth, trim=50 600 10 15, clip]{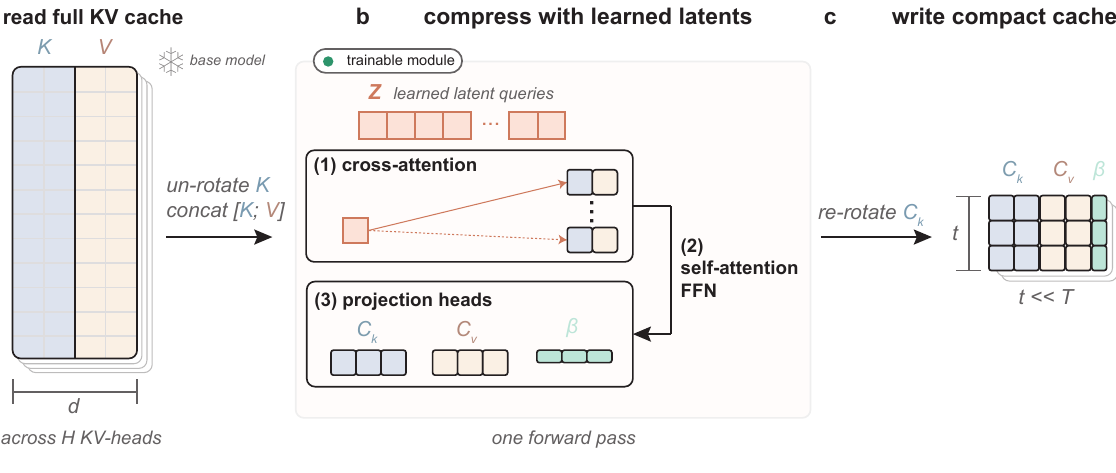}
    \caption{The Still per-layer compactor. A bank of $t$ learned
    latent queries $Z$ cross-attends the full per-layer KV cache
    (with RoPE stripped from keys), refines through self-attention
    and feed-forward sub-layers, and projects to compact keys $C_k$
    and values $C_v$ in a single forward pass. An optional bias head
    $\beta$ is shown for completeness; it is disabled ($\beta=0$) in
    all reported Still results (Appendix~\ref{app:architecture}).}
    \label{fig:diagram}
\end{figure}
 
KV cache compaction has emerged as the central primitive for managing
this gap, with existing methods falling along two axes: the compact
cache is either \emph{selected} from original tokens or
\emph{synthesized} via content-dependent combination, and the work is
performed \emph{per-context} at inference, \emph{amortized} once
offline, or \emph{trained-in} as part of pretraining. Per-context selection \citep{zhang2023h2o, li2024snapkv, cai2025pyramidkv, kim2025kvzip} scores tokens on the fly and evicts the rest, which is lightweight but structurally subset-bound. Per-context synthesis \citep{eyuboglu2025cartridges, zweiger2026fast} lifts this ceiling but requires optimization for each context. Amortized selection appeared recently in KV-Distill \citep{chari2025kvdistill}, which produces a subset of original tokens in a single prefill via a learned scorer plus LoRA \citep{hu2022lora} adaptors.

We study the underexplored combination of \emph{amortized synthesis of layer-wise KV caches for frozen pretrained models}. This targets plug-in compaction of existing open-weight checkpoints, with synthesis avoiding the subset problem of token selection and amortization avoiding per-context optimization. Representation learning reaches the same conclusion in analogous settings: amortized variational inference \citep{kingma2014auto, rezende2014stochastic} and sparse autoencoders \citep{makhzani2013ksparse, bricken2023monosemanticity} both replace repeated per-instance optimization over a fixed structure with a learned encoder. In each case, the lesson has been to learn the optimizer. We apply the same move to KV-cache synthesis.
 
We introduce \emph{Still}, a small Perceiver-based
\citep{jaegle2021perceiver, alayrac2022flamingo} compactor for this setting (Figure~\ref{fig:diagram}). Still consists of one module per transformer layer; each maintains a bank of learned latent queries that cross-attends the full KV cache and produce compact keys and values in a single forward pass, all without reference queries, gradient steps, or closed-form solves at inference. Empirically, we evaluate Still as a compaction primitive across four regimes: the single-pass speed--quality frontier across compression ratios and contexts (\S\ref{subsec:design-space}), transfer of these traits across model scales and attention architectures
(\S\ref{subsec:scaling}), long-context evidence preservation against amortized selection together with free-form summarization
(\S\ref{subsec:long-context}, \S\ref{subsec:generation}), and
iterative chunked compaction at a fixed compression ratio
(\S\ref{subsec:iterative}).

\begin{figure}
    \centering
    \includegraphics[width=0.9\linewidth]{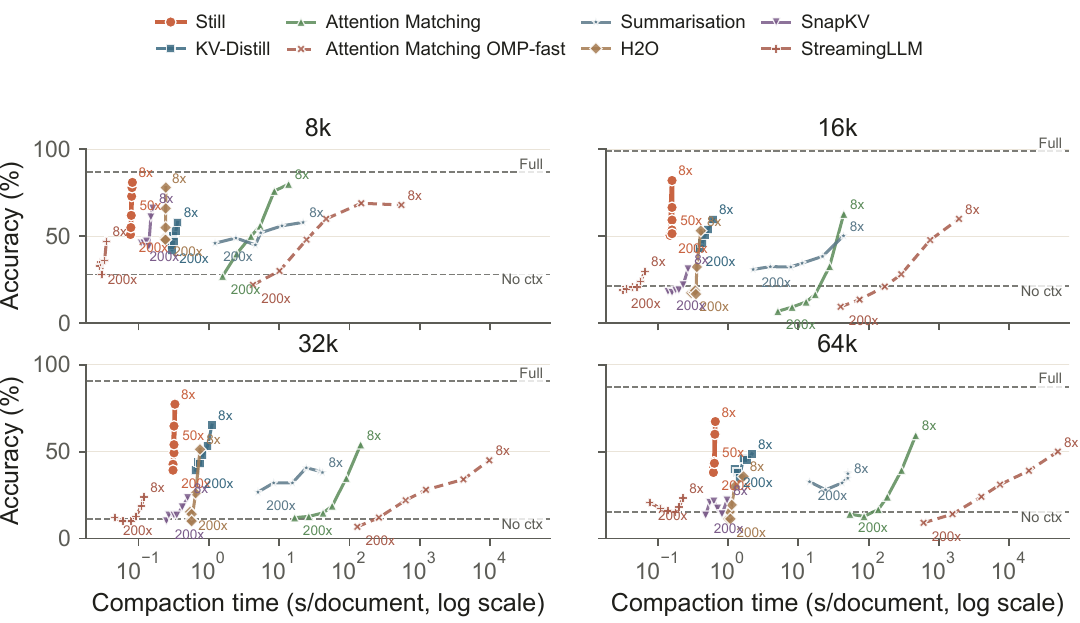}
    \caption{Speed--quality trade-off for cache compaction across
    context lengths. Each panel plots compact-cache accuracy against
    compaction time per document on a log scale, with compression
    ratios annotated along each method curve. Still stays on the fast
    side of the Pareto frontier while retaining substantially higher
    accuracy than selection-based methods at high compression,
    especially as context length increases. Summarization is a prompted
    natural-language baseline. Each Still and KV-Distill point is the
    mean over $3$ training seeds; deterministic baselines are a single
    run. KVZip is included as a targeted $25$-document partial check at
    $8$k and $16$k; see Appendix~\ref{app:kvzip-partial}. Per-cell SDs
    are reported in Table~\ref{tab:exact-results-summary}.}
    \label{fig:pareto-frontier}
\end{figure}

In the sweep on Qwen3-4B \citep{qwen2025qwen3}, Still occupies the favorable side of the speed--quality frontier across $8\times$--$200\times$ compression and $8$k--$64$k contexts (Figure~\ref{fig:pareto-frontier}). Comparable methods that are
accurate at low compression are slow, and methods that are fast lose accuracy under tight budgets; Still is the only evaluated method that remains both accurate and fast across the matrix. The same recipe transfers across Qwen3 dense models from 4B to 32B, the Qwen3-30B-A3B MoE, and Gemma-3's \citep{gemma2025technical} mixed sliding-window/global attention stack.

At long context, the matched-training RULER \citep{hsieh2024ruler} Still exceeds KV-Distill by $8$--$22$ accuracy points across
16 of 18 matched-training cells. The compact cache also supports
free-form summarization. On HELMET \citep{yen2025helmet}
\texttt{multi\_lexsum} \citep{shen2022multilexsum}, Still preserves most
of the full-context summarization gain across $8$k--$64$k contexts while
compressing the document prefix before generation, and remains ahead of
Attention Matching and KV-Distill at every length. On LongBench v1
\citep{bai2024longbench} GovReport and QMSum, Still also wins $60\%$ of pairwise judge comparisons against
KV-Distill (\S\ref{subsec:generation}). Because Still is a forward pass, it can be called repeatedly as new context arrives, and we show that this enables iterative chunked compaction at a fixed compression ratio. This is the setting in which
per-context synthesis is least practical, since its fitting cost
compounds with every compaction event. 

Together, Still is a learned primitive that uses amortization to create a compact cache that across several axes may be usable at inference time.

\section{Method}
\label{sec:method}

Let $f_\theta$ be a frozen autoregressive transformer with $L$ layers, $H$
KV-heads per layer, head dimension $d$, and rotary position embeddings
\citep{su2023roformerenhancedtransformerrotary}. Given a prefix of $T$
tokens, $f_\theta$ produces a per-layer KV cache; for layer $\ell$ and
KV-head $h$, write the cached keys and values as $K^{(\ell,h)},
V^{(\ell,h)} \in \mathbb{R}^{T \times d}$. Still learns a per-layer
compactor
\[
g_\phi^{(\ell)} :
    (K^{(\ell,h)}, V^{(\ell,h)})_{h=1}^{H}
    \mapsto
    (C_k^{(\ell,h)}, C_v^{(\ell,h)})_{h=1}^{H},
\]
with $C_k^{(\ell,h)}, C_v^{(\ell,h)} \in \mathbb{R}^{t \times d}$,
and $t \ll T$. The compact cache replaces the original prefix cache when
running $f_\theta$ on subsequent tokens. The implementation also supports the optional
log-space bias channel used by Attention Matching
\citep{zweiger2026fast}, but that channel is disabled for the reported
Still results. The base model's weights are unchanged. We describe a single layer below and suppress the layer index where unambiguous.

\subsection{Per-layer Perceiver compactor}
\label{sec:method-architecture}

Still uses one Perceiver-style compactor per transformer layer
\citep{jaegle2021perceiver, alayrac2022flamingo}. Within a layer, the compactor operates on all $H$ KV-heads jointly: each head has its own
bank of $t$ learned latent queries $Z \in \mathbb{R}^{H \times t \times
d_\ell}$, while linear projections are shared across heads.

The input to a layer's compactor is the per-head concatenation
$X^{(h)} = [K^{(h)}; V^{(h)}] \in \mathbb{R}^{T \times 2d}$, with cached keys first un-rotated into a position-free frame
(\S\ref{sec:method-pipeline}). The latents pass through $B$ pre-norm blocks; a block applies cross-attention from the latents into $X$ ie., $Z'   = Z + \mathrm{CrossAttn}(\mathrm{RMSNorm}(Z), X)$,
latent self-attention $Z''  = Z' + \mathrm{SelfAttn}(\mathrm{RMSNorm}(Z'))$, and an optional position-wise feed-forward
sublayer $Z''' = Z'' + \mathrm{FFN}(\mathrm{RMSNorm}(Z''))$ (norms are RMSNorm \citep{zhang2019rootmeansquarelayer}). Two independent linear heads, shared across KV-heads within a layer, project the final latent state $Z_\mathrm{out}$ to compact keys $C_k = Z_\mathrm{out} W_\mathrm{key}$, compact values
$C_v = Z_\mathrm{out} W_\mathrm{val}$. An optional third head can
produce per-token biases $\beta = Z_\mathrm{out} w_\beta$, but reported experiments set $\beta=0$. The only per-KV-head parameters are the latent banks. For global-attention layers the compact $(C_k, C_v)$ replace the full prefix cache; for architectures with mixed sliding-window and global attention (e.g.\ Gemma-3), Still applies
compaction only to global-attention layers.

\paragraph{Canonical configuration.}
Throughout the paper we hold the Still architecture fixed and vary only
the compact length $t$. The canonical configuration on Qwen3-4B uses
$d_\ell = 256$ and $B = 2$ blocks with cross-attention repeated in
every block; at $t = 128$ this is roughly $50$M parameters, around
$1\%$ of the base model.\footnote{Architecture sweeps
(\S\ref{app:arch-ablations}) show that headline accuracy is
insensitive to most architectural choices; the binding constraint is
within-layer coordination, not capacity.} Full hyperparameters
($n_c, n_s$ heads, FFN setting, QK-norm and logit scale, RoPE base,
and optional $\beta$ support) and the detailed cross-attention math are in
Appendix~\ref{app:architecture}.

\subsection{Position-free compaction}
\label{sec:method-pipeline}

The base model's cached keys are already RoPE-rotated \citep{su2023roformerenhancedtransformerrotary}, so the same
content has different key vectors at different positions. Forcing the
compactor to synthesize mixtures of already-rotated keys mixes content
with phase and is empirically unstable. Still therefore operates in a
position-free frame: cached keys are inverse-rotated before the
compactor, the compactor uses its own RoPE inside cross-attention with
latent queries placed at evenly spaced positions over the compacted
range, and the resulting compact keys are re-rotated at chosen output
positions before being written into the cache. Values are not rotated. We use an identity-style initialization that makes the compactor a
near-pass-through map at $t = T$; it
requires $d_\ell = 2d$ and is described in
Appendix~\ref{app:identity-init}.

\subsection{Single-pass and iterative compaction}
\label{sec:method-iterative}

\paragraph{Single-pass compaction.}
For a static prefix, Still runs the frozen base model once to obtain
the full per-layer KV cache, applies the per-layer compactors to that
cache, and substitutes the resulting $(C_k, C_v)$ for the
original prefix entries. The continuation path (question, generation,
or downstream tokens) is processed against the compact prefix
unchanged, with continuation positions offset so that RoPE phases on
new tokens behave as if the full original prefix were still present.

\paragraph{Iterative chunked compaction.}
For long-horizon settings where context arrives over time, Still
supports a recurrent schedule with fixed local compression ratio $c$. Pass 0 prefills the first $2ct$ tokens and then compacts the first $ct$ KV entries to $t$ entries. Each subsequent pass $N>0$ prefills the next outstanding chunk of $ct$ tokens, conditioned on the $N$ already compacted KV chunks concatenated with a raw KV chunk. In other words, the base model at pass $N$ sees a KV cache of length $(N+c)t$, and uses this to prefill the next $ct$ tokens. The chunk of $ct$ tokens preceding this newly prefilled one is then compacted to $t$ entries. The retained cache after ingesting $T$ tokens is therefore $T/c + ct$ entries: linear in context at compression rate $1/c$, rather than constant. This is a fixed local compression ratio with bounded per-chunk growth, not a constant absolute budget; we discuss truly constant-budget recurrences as an open direction in \S\ref{sec:discussion}.
 
It is crucial for prefill to occur sequentially this way in tandem with compaction because this matches the use case at inference over long contexts. We provide a one chunk buffer of raw KV cache at every prefill because it prevents infidelity of the compaction from compounding too aggressively. We refer to this buffering mechanism as lookahead. Compact-key re-rotation, position-offset accounting, and the chunked-prefill / lookahead pipeline used as the default for iterative results are in
Appendix~\ref{app:pipeline}.

\subsection{Training}
\label{sec:method-training}

\paragraph{Data.}
The canonical training corpus is a four-domain extractive
multiple-choice QA dataset spanning Financial filings, Project
Gutenberg literature, Legal documents, and Code, with each example a
triple of (long-context prefill, MCQ prompt, answer-letter target).
Questions are generated from random sub-chunks of the prefill,
verified by the same frozen base model used as teacher, and filtered
to remove items the base model can answer without the context. Domains
are downsampled to the smallest, giving $\sim 120$k items and $\sim 1$B
context tokens at $8$k context length; $16$k single-pass and
$\{8\text{k}, 16\text{k}, 32\text{k}\}$ iterative families use matched
splits prepared the same way. Per-domain item counts are in
Appendix~\ref{app:training}.

\paragraph{Training.}
The training signal is forward KL divergence from a full-context
teacher to the compact-cache student, masked to answer-side tokens
only:
\[
\mathcal{L}
=
\mathbb{E}_{(c, p, a) \sim \mathcal{D}}
\sum_{i \in \mathrm{ans}}
\mathrm{KL}\!\left(
    f_\theta(\cdot \mid c, p, a_{<i})
    \,\|\,
    f_\theta(\cdot \mid g_\phi(c), p, a_{<i})
\right),
\]
where teacher and student are the same frozen base model differing only
in whether the prefix cache is the full uncompressed cache or Still's
compact cache. Answer tokens consist of a rationale trace followed by the MCQ answer token to provide a richer feedback signal than what would be available from just the single MCQ answer token. The KL is evaluated on the top $200$ teacher-vocabulary
tokens with the gold answer token forced into the support, which keeps
the loss memory-bounded; Appendix~\ref{app:kl-support-ablation} validates
this choice by finding no gain from top-$1000$ or full-vocabulary support
under a common full-vocabulary evaluation. The compactor parameters $\phi$ are the
only trainable parameters; $\theta$ remains frozen throughout. Cache-MSE
and residual variants are ablations only
(Appendix~\ref{app:factorization-ablation}). Except for the explicitly
labeled generation-aligned LongBench summarization checkpoint in
\S\ref{subsec:generation}, paper-facing Still checkpoints use pure KL.

We train with AdamW \citep{loshchilov2019adamw} at learning rate $4 \times 10^{-5}$ on $8 \times$
H200 with DDP at effective batch $32$, taking all
Qwen3-4B checkpoints at training step $1500$. Full hyperparameters
(weight decay, warmup, schedule, gradient clipping) are in
Appendix~\ref{app:training}.

We report mean (SD) over $3$ training seeds for all frontier
results (Table~\ref{tab:exact-results-summary},
Figure~\ref{fig:compression-sweep}, Figure~\ref{fig:pareto-frontier},
Figure~\ref{fig:still-delta-panels}).
Deterministic baselines (H2O,
SnapKV, StreamingLLM, Attention Matching, AM-OMP-fast) are
seed-invariant; their reported SD is bootstrap over evaluation
examples. The RULER cells (\S\ref{subsec:long-context}) and the
RULER and LongBench v2 rows of the Still-vs-KV-Distill summary
(Table~\ref{tab:still-kvdistill-summary}) are reported as
3-seed means with seed-level SDs. HELMET
(\S\ref{subsec:generation}), iterative-compaction
(\S\ref{subsec:iterative}), and the QuALITY-derived row of
Table~\ref{tab:still-kvdistill-summary} remain single-seed with
bootstrap-over-examples SDs, and we flag this in those sections.

\subsection{Baselines}
\label{sec:method-baselines}

We compare Still against representative methods from the populated
regions of this design space.
\emph{Per-context selection}: H2O \citep{zhang2023h2o} retains tokens
by cumulative attention scores from extracted reference queries;
SnapKV \citep{li2024snapkv} scores tokens within a $64$-token
observation window with a $5$-token pooling kernel; StreamingLLM
\citep{xiao2024efficient} keeps $4$ attention-sink tokens plus the
most recent $K - 4$ tokens. \emph{Per-context synthesis}: Attention
Matching \citep{zweiger2026fast} uses repeat-prefill query extraction,
top-$k$ key selection by RMS score, least-squares value
reconstruction, and nonnegative least-squares $\beta$ fitting.
\emph{Amortized selection}: KV-Distill \citep{chari2025kvdistill} uses a learned token scorer with
top-$k$ retention, rank-$128$ LoRA adaptors on $Q, K, V, O$, and
forced sink tokens with an uncompressed question/answer path. Each
baseline runs a single configuration across all evaluations, with only
the cache budget $K$ varying to match Still's compression ratio.
Implementation details, including the KV-Distill retraining protocol
and Attention Matching's reference-query budget, are in
Appendix~\ref{app:baselines}.

\section{Results}
\label{sec:results}
 
Each subsection asks where Still sits relative to populated alternatives in the cache-compaction design space. We report the speed--quality frontier across compression ratios and context lengths (\S\ref{subsec:design-space}), transfer to other model scales and to a mixed sliding-window/global attention stack (\S\ref{subsec:scaling}), long-context behavior against the strongest amortized-selection baseline (\S\ref{subsec:long-context}), iterative compaction at a fixed compression ratio (\S\ref{subsec:iterative}), and free-form generation (\S\ref{subsec:generation}).
 
% =====================================================================
\subsection{Amortized synthesis offers the best speed--quality tradeoff across the design space}
\label{subsec:design-space}
 
We sweep compression ratios from $8\times$ to $200\times$ across ${8\text{k},16\text{k},32\text{k},64\text{k}}$ contexts, comparing Still with per-context selection methods (H2O, SnapKV, StreamingLLM), per-context synthesis (Attention Matching), amortized selection (KV-Distill), and a prompted natural-language summarization baseline. The main result is the frontier shape (Figure~\ref{fig:pareto-frontier}): selection methods degrade under tight budgets, prompted summaries preserve some question-relevant content but pass through a lossy text bottleneck, and Still is the only method that remains accurate across the sweep. We also ran a targeted partial KVZip comparison at $8$k and $16$k because it is the closest recent query-agnostic neighbor; Appendix~\ref{app:kvzip-partial} reports the results and why we do not treat that reference-port check as a full-matrix baseline. A 256k-token single-pass Still compactor at 100× compression reaches 40.7\% (3-seed $\sigma=3.2$) compact accuracy on a QuALITY-concat diagnostic, with full-context (55\%, $\sigma=1.8$) and no-context (22\%, $\sigma=1.5$), showing the same mechanism can be trained directly at much longer contexts. All frontier numbers in Figure~\ref{fig:pareto-frontier}, Table~\ref{tab:exact-results-summary}, and Figure~\ref{fig:still-delta-panels} are means over $3$ training seeds for the learned methods (Still, KV-Distill); RULER (\S\ref{subsec:long-context}) is also reported as 3-seed means with seed-level SDs, while HELMET (\S\ref{subsec:generation}) and iterative compaction (\S\ref{subsec:iterative}) are reported single-seed with bootstrap-over-examples SDs.
 
% =====================================================================
\subsection{The same compactor transfers across model scales and attention architectures}
\label{subsec:scaling}
 
\begin{figure}
\centering
\begin{minipage}[t]{0.55\linewidth}
\centering
\includegraphics[width=\linewidth]{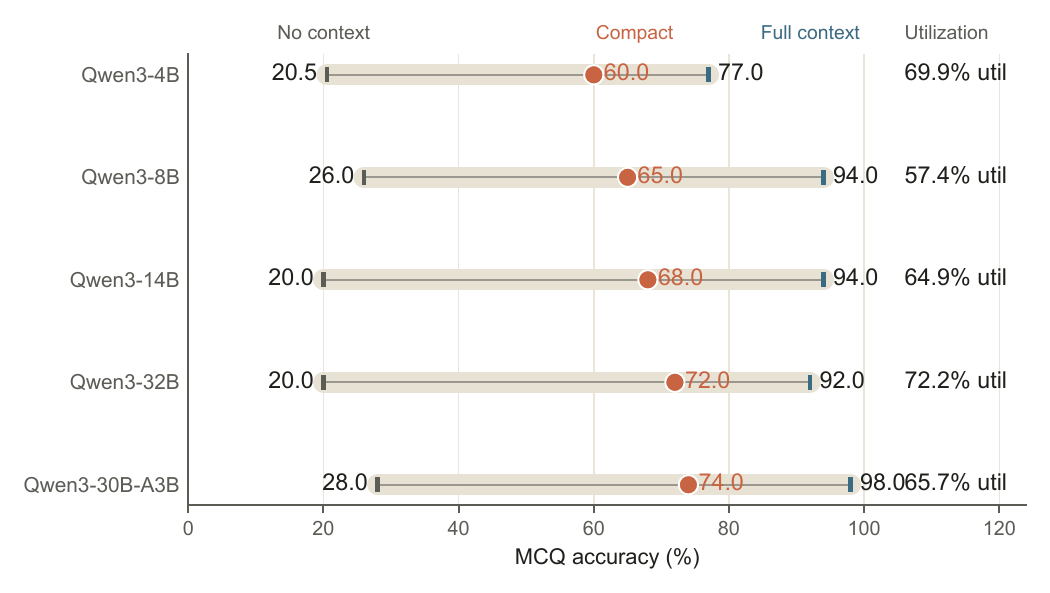}
\caption{Qwen3 dense models (4B, 8B, 14B, 32B) and the Qwen3-30B-A3B MoE at fixed $50\times$ compression.}
\label{fig:scale-moe}
\end{minipage}\hfill
\begin{minipage}[t]{0.43\linewidth}
\centering
\includegraphics[width=\linewidth]{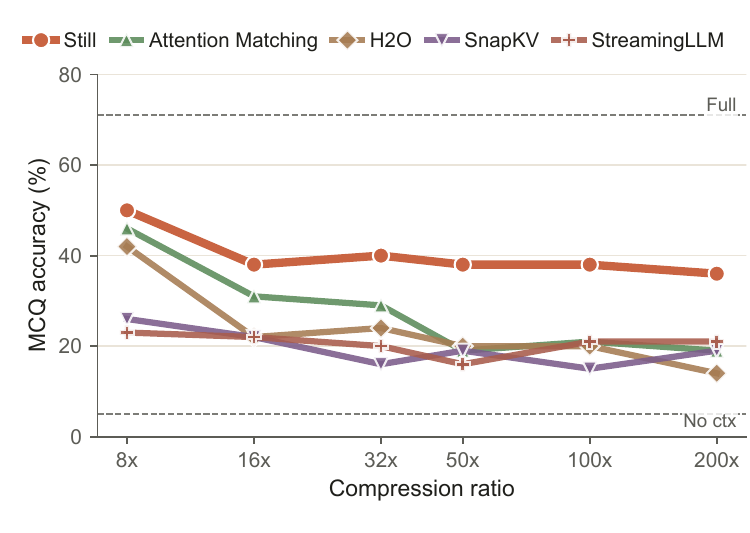}
\caption{Still vs.\ fixed-cache baselines on Gemma-3 4B at 16k context.}
\label{fig:gemma3-fixed}
\end{minipage}
\end{figure}
 
Holding architecture, training corpus, optimizer, schedule, and compactor depth fixed, Figure~\ref{fig:scale-moe} reports compact-cache utility across the Qwen3 dense family from 4B to 32B with the 30B-A3B mixture-of-experts
on the same plot (left), and on Gemma-3 4B's mixed sliding-window/global attention stack (right). Across the Qwen3 dense models the compact cache stays usefully inside the no-context-to-full-context band at every size; the MoE transfers without modification, since the compact-cache interface is determined by attention-layer KV geometry and FFN structure is not part of that interface. Gemma-3 transfers under a single specialization i.e., compact only the global-attention layers, leaving sliding-window layers (which already have a bounded cache) untouched with $\beta = 0$. This recipe matches or exceeds every fixed-cache baseline on Gemma-3 4B at 16k.
 
% =====================================================================
\subsection{Long-context generalization against amortized selection}
\label{subsec:long-context}
 
\begin{figure}
\centering
\begin{subfigure}[t]{0.49\linewidth}
\centering
\includegraphics[width=\linewidth]{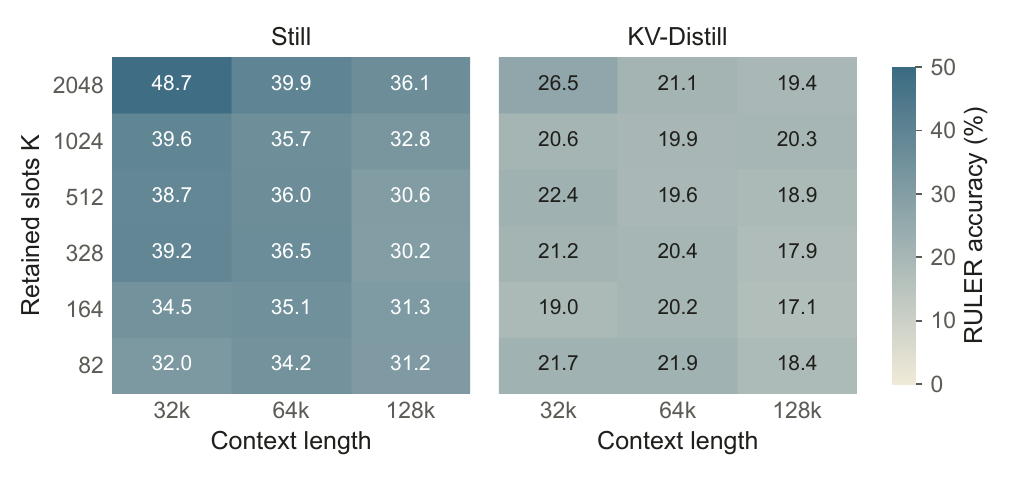}
\caption{Matched-training holdout. Compactors trained directly on RULER tasks, evaluated on held-out RULER tasks at 32k/64k/128k.}
\label{fig:ruler-matched}
\end{subfigure}\hfill
\begin{subfigure}[t]{0.49\linewidth}
\centering
\includegraphics[width=\linewidth]{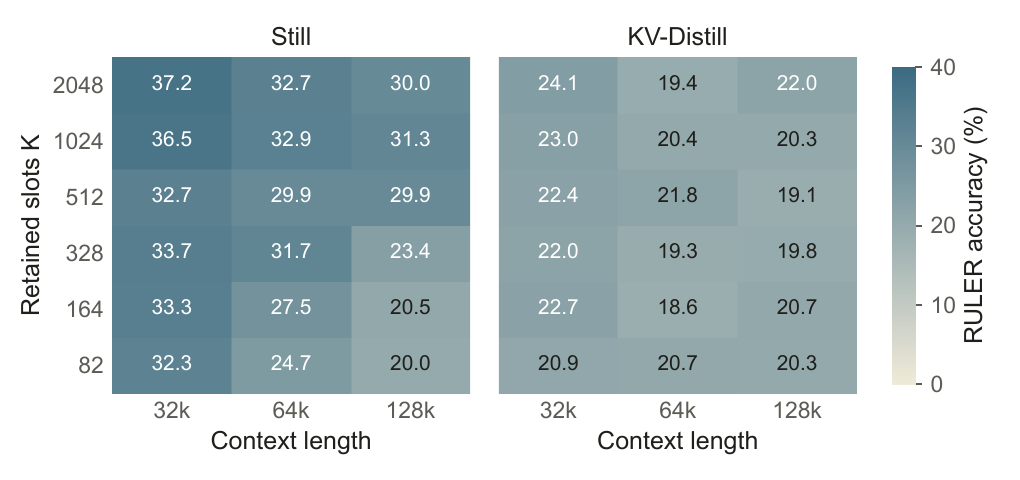}
\caption{Zero-shot transfer. Compactors trained on the mixed-domain MCQ corpus, evaluated on RULER without retraining.}
\label{fig:ruler-transfer}
\end{subfigure}
\caption{Still vs.\ KV-Distill on RULER at long context. Each cell reports compact-cache accuracy at a fixed final cache size $K$, with a shared color scale across the two methods. Per-cell 3-seed means and seed-level SDs are in Table~\ref{tab:ruler-cells}; cells where the Still--KV-Distill gap falls within seed-level noise (gap/SE $<$ 2) are noted there.}
\label{fig:ruler-combined}
\end{figure}
 
The cleanest comparison between amortized synthesis and amortized selection is at matched training and matched evaluation. Figure~\ref{fig:ruler-matched} reports compactors trained directly on RULER tasks and evaluated on held-out RULER tasks at 32k/64k/128k; Figure~\ref{fig:ruler-transfer} reports the more demanding transfer setting from the mixed-domain MCQ corpus. We include LongBench v2 \citep{bai2024longbenchv2} (a 4-way multiple-choice benchmark, distinct from the LongBench v1 free-form summarization audit in \S\ref{subsec:generation}) in Table~\ref{tab:still-kvdistill-summary} as an additional transfer diagnostic.
 
Under matched training, Still exceeds KV-Distill by $8$--$22$ accuracy points in 16 of 18 cells. The two exceptions (64k/$K{=}82$ and 128k/$K{=}82$) sit within seed-level noise, isolating the failure regime as long context combined with extreme compression. The transfer setting tells the same qualitative story with smaller margins; the within-noise region extends to a handful of additional cells at $K \le 164$ on 128k as the train--test mismatch tax compounds with proximity to the no-context floor. Per-cell 3-seed means and seed-level SDs are in Table~\ref{tab:ruler-cells}.
 
% =====================================================================
\subsection{Iterative compaction at fixed compression ratio}
\label{subsec:iterative}
 
\begin{figure}
\centering
\includegraphics[width=0.75\linewidth]{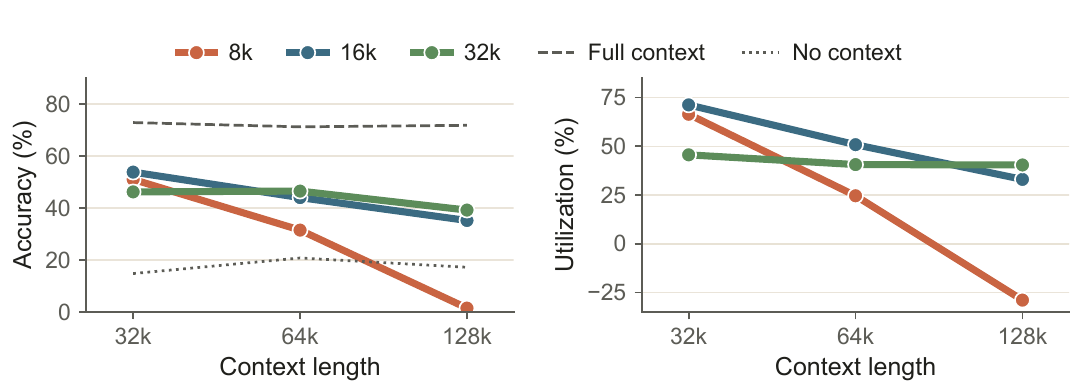}
\caption{Iterative compaction extends the usable context window on Long-MCQ. Left: compact-cache accuracy with full-context and no-context references. Right: normalized utilization.}
\label{fig:iter-mudith-main}
\end{figure}
 
In long-horizon settings, context arrives over time and the deployment constraint is a fixed compression ratio: the retained cache grows with the trajectory at rate $1/c$ rather than at full token rate, with bounded per-chunk overhead. Per-context synthesis methods are operationally unattractive in this regime because their per-invocation optimization compounds across the trajectory. With Still, mid-trajectory compaction is a forward pass. All iterative results in this section use $\beta=0$.

Figure~\ref{fig:iter-mudith-main} evaluates iterative chunked compaction (\S\ref{sec:method-iterative}) on Long-MCQ, comparing Still checkpoints trained at three iterative horizons (8k, 16k, 32k) deployed at 32k, 64k, and 128k. Each checkpoint produces $t = 128$ latents per chunk at a fixed per-chunk compression ratio $c$ tied to its training horizon: $1024$-token chunks at $c = 8$ (8k-trained), $2048$-token chunks at $c = 16$ (16k-trained), and $4096$-token chunks at $c = 32$ (32k-trained). All cells use the lookahead variant (one-chunk raw-KV buffer, Appendix~\ref{app:pipeline}); the number of stacked compaction passes ranges from $8$ (32k-trained at 32k) to $128$ (8k-trained at 128k). The 8k-trained compactor reaches $51.0\%$ at 32k but collapses to $1.5\%$ at 128k, below the no-context floor; training on longer iterative contexts removes this collapse, with the 16k-trained checkpoint reaching $35.2\%$ and the 32k-trained checkpoint $39.2\%$ at 128k, with the same pattern reflected in utilization. The collapse is not a cache-budget effect: by the recurrence accounting in \S\ref{sec:method-iterative}, the 8k-trained checkpoint retains a \emph{larger} cache at 128k ($17408$ slots) than the 32k-trained one ($8192$); the binding constraint is the training horizon, since the 8k checkpoint never saw $128$ stacked compactions during training. Additional dataset and task diagnostics, including QuALITY and RULER iterative sweeps and the evaluation-mode ablation, are in Appendix~\ref{app:iterative-compaction}.
 
% Figure~\ref{fig:iter-mudith-main} evaluates iterative chunked compaction (\S\ref{sec:method-iterative}) on Long-MCQ, comparing Still checkpoints trained at three iterative horizons (8k, 16k, 32k) deployed at 32k, 64k, and 128k. The 8k-trained compactor reaches $51.0\%$ at 32k but collapses to $1.5\%$ at 128k, below the no-context floor; training on longer iterative contexts removes this collapse, with the 16k-trained checkpoint reaching $35.2\%$ and the 32k-trained checkpoint $39.2\%$ at 128k, with the same pattern reflected in utilization. \textcolor{blue}{We need to specify chunk size and compaction ratio - are all of these lookahead?} However, iterative compaction is not free extrapolation: training horizon is the major factor governing where the recurrence breaks down, but absolute deployment length still matters, with the 32k-trained model degrading from 32k to 128k just much more gracefully than the 8k-trained one. Additional dataset and task diagnostics, including QuALITY and RULER iterative sweeps and the evaluation-mode ablation, are in Appendix~\ref{app:iterative-compaction}.

% =====================================================================
\subsection{Free-form summarization preserves most of the full-context gain}
\label{subsec:generation}

\begin{table}[h]
\centering
\begin{minipage}[t]{0.49\textwidth}
\centering
\scriptsize
\captionof{table}{HELMET \texttt{multi\_lexsum} free-form summarization utilization.}
\label{tab:helmet-multi-lexsum}
\resizebox{\linewidth}{!}{%
\begin{tabular}{lccc}
\toprule
Context & Still & AM & KV-Distill \\
\midrule
8k  & \textbf{94.9 (1.4)} & 88.0 (1.6) & 85.5 (1.7) \\
16k & \textbf{86.8 (1.7)} & 80.3 (1.5) & 76.0 (1.9) \\
32k & \textbf{81.3 (1.5)} & 72.1 (1.8) & 68.8 (1.6) \\
64k & \textbf{74.0 (1.8)} & 65.2 (1.4) & 59.9 (1.5) \\
128k & \textbf{59.3 (1.6)} & 49.9 (1.7) & 43.5 (1.8) \\
\bottomrule
\end{tabular}%
}
\end{minipage}
\hfill
\begin{minipage}[t]{0.49\textwidth}
\centering
\scriptsize
\captionof{table}{LongBench free-form summarization at 16k context and $K=1024$.}
\label{tab:longbench-generation-pairwise}
\resizebox{\linewidth}{!}{%
\begin{tabular}{lccc}
\toprule
Task & Still (gen.) & KV-Distill (gen.) & $\Delta$ \\
\midrule
Overall & \textbf{58.8} & 53.4 & \textbf{+5.4} \\
GovReport & \textbf{55.6} & 48.6 & \textbf{+7.0} \\
QMSum & \textbf{62.1} & 58.2 & \textbf{+3.8} \\
\bottomrule
\end{tabular}%
}
\end{minipage}
\end{table}

HELMET summarization asks whether the compact cache can support free-form summarization rather than only answer selection. Table~\ref{tab:helmet-multi-lexsum} reports Still, Attention Matching (AM), and KV-Distill on \texttt{multi\_lexsum}, where each summary is scored by HELMET's GPT-4o judge against full-context and no-context references. Across 8k--64k contexts, Still recovers $74$--$95\%$ of the full-context gain while compressing the document prefix before generation, and it still recovers $59\%$ at 128k. AM and KV-Distill both lag Still at every context length, with the gap to KV-Distill widening from $\sim 9$ points at 8k to nearly $16$ points by 128k, the same ordering observed on matched-training RULER (\S\ref{subsec:long-context}).

LongBench v1 \citep{bai2024longbench} (open-ended summarization; distinct from the LongBench v2 multiple-choice transfer diagnostic in \S\ref{subsec:long-context}) asks for direct open-ended summarization rather than normalized utilization against full- and no-context anchors. Table~\ref{tab:longbench-generation-pairwise} reports a held-out GovReport/QMSum comparison between Still and KV-Distill at the same 16k context and $K=1024$ retained-cache budget. Both checkpoints are short generation-aligned continuations of their respective 16k/$K{=}1024$ checkpoints, trained on GovReport/QMSum summary rows under the identical recipe; sampling, training, and judge details are in Appendix~\ref{app:longbench-generation-method}. Still wins $300$ of $500$ pairwise comparisons and improves the mean GPT-5.5 judge score by $5.4$ points.

\section{Related Work}
\label{sec:related}

The KV cache compaction literature~\citep{kwon2023efficient} spans selection and synthesis methods with per-context, amortized, or trained-in computation. Per-context selection methods~\citep{zhang2023h2o, li2024snapkv, cai2025pyramidkv, xiao2024efficient, kim2025kvzip} score tokens at inference and evict the rest; the compact entries must be rows of the original cache, so the cell's quality ceiling falls with the input's information density. The remaining regions are populated more sparsely, and we compare Still against a representative method from each.

KV-Distill~\citep{chari2025kvdistill} is the closest published neighbor: a learned scorer identifies tokens to retain via top-$k$, and rank-$128$ LoRA adaptors on the base model's $Q, K, V, O$ matrices route retained tokens through trainable projections during a modified prefill, trained with symmetric KL against the uncompressed-cache teacher. Each compact entry remains a row of the original cache; the LoRA paths modulate how attention reads the cache rather than what it contains. Still inherits the amortization move but breaks the selection constraint --- a single Still slot can blend information from arbitrarily many positions via cross-attention --- and leaves the base model's weights unchanged. The cell-level gap (\S\ref{subsec:long-context}) widens with both context length and compression ratio.

The strongest quality references in the literature compact in latent space rather than by selection. Cartridges~\citep{eyuboglu2025cartridges} treats the compact cache as a parameter and trains it via prefix tuning~\citep{li2021prefix} on synthetic ``self-study'' interactions generated by the base model; structural analysis of trained Cartridges~\citep{routers2025cartridges} finds that the keys act as stable retrieval routers while values carry most of the compressed content. Attention Matching~\citep{zweiger2026fast} preserves attention outputs and per-head attention mass via closed-form key selection, least-squares value reconstruction, and nonnegative least-squares fitting of per-token biases. Still inherits the latent-space synthesis paradigm from this line, supports the same optional bias interface for diagnostics, and removes per-context optimization from the reported method: compaction is a single forward pass through a learned module, which is the operational distinction that enables iterative long-horizon compaction (\S\ref{subsec:iterative}).

A separate axis reduces cache cost by changing the model itself, from grouped-query attention~\citep{ainslie2023gqa} and Dynamic Memory Compression~\citep{nawrot2024dynamic} to multi-head latent attention~\citep{deepseekai2024v3} and native sparse attention~\citep{yuan2025nsa}. The most sophisticated synthesis-based instance is DeepSeek-V4~\citep{deepseekai2026deepseekv4}, which interleaves softmax-gated pooling, sparse top-$k$ block selection, and a sliding-window path. Still and V4 both synthesize compact entries rather than selecting original tokens, but V4 bakes compression into pretraining over a fixed local receptive field, whereas Still trains a small module against a frozen checkpoint with global cross-attention; the convergence on synthesis is independent evidence for the design choice.

Still's architecture is a small Perceiver~\citep{jaegle2021perceiver}, closest in application to the Perceiver Resampler in Flamingo~\citep{alayrac2022flamingo}, which distils variable-length visual features into a fixed token budget for language-model conditioning. Earlier context-compression work~\citep{mu2023gisting, chevalier2023adapting, ge2023incontextautoencoder} uses related token-bottleneck ideas but operates in input or activation space rather than producing layer-wise compact KV caches.

\section{Discussion}
\label{sec:discussion}

The iterative-compaction result in \S\ref{subsec:iterative} cuts both ways. The 16k- and 32k-trained compactors absorb context many times their physical capacity at 128k, but the 8k-trained compactor collapses below the no-context floor and the 32k-trained compactor still degrades meaningfully from 32k to 128k. Iterative compaction is not free extrapolation; the failure is graceful only within the trained range. Supporting deployment at, say, 1M tokens therefore requires training at iterative horizons proportionally longer than 16k or 32k, which is expensive both in token count and in the difficulty of curating long-context training data. Curriculum schedules over horizon, training Still directly under the recurrence, and constant-budget variants that reuse or merge compact slots in place are all natural follow-ups. The optional $\beta$-scaling diagnostic in Appendix~\ref{app:arch-ablations} illustrates why this matters: post-hoc calibration that looks reasonable in single-pass compaction can become miscalibrated under recurrence, even though the reported Still results avoid that deployment knob by setting $\beta=0$. Once compaction itself is cheap, the binding constraint shifts to training a compactor whose distribution covers the lengths and intermediate cache states it will encounter at deployment.

HELMET \texttt{multi\_lexsum} and LongBench v1 summarization (\S\ref{subsec:generation}) extend this from answer selection to free-form summarization. The compact cache is a usable conditioning state for the frozen base model in both retrieving evidence under multiple-choice scoring and for generating summaries from a compressed document prefix. The comparison to KV-Distill is precisely useful because both methods are amortized, but KV-Distill can only retain original cache rows whereas Still can synthesize compact entries. This supports an interpretation of a compressed working memory that the model can attend to natively, without lossless recall of every original token. Critically, this does not establish arbitrary open-ended generation or exact recall.

The matched-training RULER result gives the cleanest comparison, as we fix the training distribution and change the compaction cell from amortized selection to amortized synthesis. The transfer cells ask a harder question, whether the learned compact state remains useful when the task moves away from training. The smaller margins we observe there, and the occasional approach to parity (Table~\ref{tab:still-kvdistill-summary}), are therefore best read as a train--test mismatch rather than a reversal of the main result. Clearly, scaling Still will require training mixtures that cover the task surfaces on which the compact cache is expected to operate.

\paragraph{Limitations.} Still is not lossless: full-context inference remains substantially better in the hardest 128k settings, and iterative compaction is not free extrapolation because the training horizon strongly affects where repeated compression breaks down. The iterative recurrence as described here also holds only a fixed compression ratio rather than a constant absolute cache budget --- the retained cache grows linearly with trajectory length at rate $1/c$ --- so the long-horizon regime we report is one of bounded per-chunk growth, not $O(1)$ recurrent memory; an in-place slot-reuse variant is left to future work. The current recurrence also remains weak on exact-retrieval and needle-style tasks, even when it preserves enough semantic evidence for broader MCQ-style evaluations, and most headline results are still extractive multiple-choice rather than open-ended deployment workloads. Operationally, single-pass Still runs after a full prefill of the original prefix, so in that mode it reduces the retained cache and subsequent attention cost rather than eliminating the initial full-context pass; iterative chunked compaction
(\S\ref{sec:method-iterative}) avoids the full-prefix pass by ingesting context in chunks, but inherits the training-horizon constraints above. Each base checkpoint also needs its own trained compactor, adding parameters, training data requirements, and training cost. Finally, although our implementation supports an optional per-token bias channel, the reported Still results use $\beta=0$; enabling such biases in production would require attention-kernel support or a slower custom path.

\section{Conclusion}
\label{sec:conclusion}

Still is a learned KV-state transform that is lightweight enough to call repeatedly, expressive enough to beat token selection when information density is high, stable enough to transfer across model scales and attention architectures, and useful enough to support both answer selection and free-form summarization. Amortization makes cache compaction tractable at long context; in the regimes we evaluate, synthesis gives the compact state enough expressive capacity to be worth amortizing.

\bibliographystyle{plainnat}
\bibliography{bibliography}

@article{gemma2025technical,
  title={Gemma 3 Technical Report},
  author={{Gemma Team}},
  journal={arXiv preprint arXiv:2503.19786},
  year={2025}
}

@inproceedings{kwon2023efficient,
  author    = {Kwon, Woosuk and Li, Zhuohan and Zhuang, Siyuan and Sheng, Ying and Zheng, Lianmin and Yu, Cody Hao and Gonzalez, Joseph E. and Zhang, Hao and Stoica, Ion},
  title     = {Efficient Memory Management for Large Language Model Serving with {PagedAttention}},
  booktitle = {Proceedings of the ACM SIGOPS 29th Symposium on Operating Systems Principles},
  year      = {2023}
}

@misc{zweiger2026fast,
  title         = {Fast {KV} Compaction via Attention Matching},
  author        = {Adam Zweiger and Xinghong Fu and Han Guo and Yoon Kim},
  year          = {2026},
  eprint        = {2602.16284},
  archivePrefix = {arXiv},
  primaryClass  = {cs.LG},
  url           = {https://arxiv.org/abs/2602.16284}
}

@article{eyuboglu2025cartridges,
  title   = {Cartridges: Lightweight and General-Purpose Long Context Representations via Self-Study},
  author  = {Eyuboglu, Sabri and Ehrlich, Ryan and Arora, Simran and Guha, Neel and Zinsley, Dylan and Liu, Emily and Tennien, Will and Rudra, Atri and Zou, James and Mirhoseini, Azalia and R{\'e}, Christopher},
  journal = {arXiv preprint arXiv:2506.06266},
  year    = {2025}
}

@misc{routers2025cartridges,
  title         = {Learned Structure in Cartridges: Keys as Shareable Routers in Self-Studied Representations},
  author        = {Diaz, Maurizio},
  year          = {2025},
  eprint        = {2508.17032},
  archivePrefix = {arXiv},
  primaryClass  = {cs.LG},
  url           = {https://arxiv.org/abs/2508.17032}
}

@inproceedings{zhang2023h2o,
  title     = {{H2O}: Heavy-Hitter Oracle for Efficient Generative Inference of Large Language Models},
  author    = {Zhang, Zhenyu and Sheng, Ying and Zhou, Tianyi and Chen, Tianlong and Zheng, Lianmin and Cai, Ruisi and Song, Zhao and Tian, Yuandong and R{\'e}, Christopher and Barrett, Clark and Wang, Zhangyang and Chen, Beidi},
  booktitle = {Advances in Neural Information Processing Systems},
  year      = {2023}
}

@inproceedings{li2024snapkv,
  title     = {{SnapKV}: {LLM} Knows What You Are Looking for Before Generation},
  author    = {Li, Yuhong and Huang, Yingbing and Yang, Bowen and Venkitesh, Bharat and Locatelli, Acyr and Ye, Hanchen and Cai, Tianle and Lewis, Patrick and Chen, Deming},
  booktitle = {Advances in Neural Information Processing Systems},
  year      = {2024}
}

@inproceedings{cai2025pyramidkv,
  title     = {{PyramidKV}: Dynamic {KV} Cache Compression Based on Pyramidal Information Funneling},
  author    = {Cai, Zefan and Zhang, Yichi and Gao, Bofei and Liu, Yuliang and Liu, Tianyu and Lu, Keming and Xiong, Wayne and Dong, Yue and Hu, Junyang and Xiao, Wen},
  booktitle = {Second Conference on Language Modeling},
  year      = {2025}
}

@inproceedings{kim2025kvzip,
  title     = {{KVzip}: Query-Agnostic {KV} Cache Compression with Context Reconstruction},
  author    = {Kim, Jang-Hyun and Kim, Jinuk and Kwon, Sangwoo and Lee, Jae W. and Yun, Sangdoo and Song, Hyun Oh},
  booktitle = {Advances in Neural Information Processing Systems},
  year      = {2025}
}

@misc{chari2024compactor,
  title         = {Compactor: Calibrated Query-Agnostic {KV} Cache Compression with Approximate Leverage Scores},
  author        = {Vivek Chari and Benjamin Van Durme},
  year          = {2025},
  eprint        = {2507.08143},
  archivePrefix = {arXiv},
  primaryClass  = {cs.CL}
}

@misc{chari2025kvdistill,
  title         = {{KV-Distill}: Nearly Lossless Learnable Context Compression for {LLMs}},
  author        = {Vivek Chari and Guanghui Qin and Benjamin Van Durme},
  year          = {2025},
  eprint        = {2503.10337},
  archivePrefix = {arXiv},
  primaryClass  = {cs.CL}
}

@misc{moschella2026learning,
  title         = {Learning to Evict from Key-Value Cache},
  author        = {Luca Moschella and Laura Manduchi and Ozan Sener},
  year          = {2026},
  eprint        = {2602.10238},
  archivePrefix = {arXiv},
  primaryClass  = {cs.CL}
}

@inproceedings{xiao2024efficient,
  title     = {Efficient Streaming Language Models with Attention Sinks},
  author    = {Xiao, Guangxuan and Tian, Yuandong and Chen, Beidi and Han, Song and Lewis, Mike},
  booktitle = {The Twelfth International Conference on Learning Representations},
  year      = {2024}
}

@inproceedings{ainslie2023gqa,
  title     = {{GQA}: Training Generalized Multi-Query Transformer Models from Multi-Head Checkpoints},
  author    = {Ainslie, Joshua and Lee-Thorp, James and de Jong, Michiel and Zemlyanskiy, Yury and Lebr{\'o}n, Federico and Sanghai, Sumit},
  booktitle = {Proceedings of the 2023 Conference on Empirical Methods in Natural Language Processing},
  year      = {2023}
}

@inproceedings{nawrot2024dynamic,
  title     = {Dynamic Memory Compression: Retrofitting {LLMs} for Accelerated Inference},
  author    = {Nawrot, Piotr and {\L}a{\'n}cucki, Adrian and Chochowski, Marcin and Tarjan, David and Ponti, Edoardo M.},
  booktitle = {Proceedings of the 41st International Conference on Machine Learning},
  pages     = {37396--37412},
  year      = {2024}
}

@misc{deepseekai2024v3,
  title         = {{DeepSeek-V3} Technical Report},
  author        = {{DeepSeek-AI}},
  year          = {2024},
  eprint        = {2412.19437},
  archivePrefix = {arXiv}
}

@inproceedings{yuan2025nsa,
  title     = {Native Sparse Attention: Hardware-Aligned and Natively Trainable Sparse Attention},
  author    = {Yuan, Jingyang and Gao, Huazuo and Dai, Damai and Luo, Junyu and Zhao, Liang and Zhang, Zhengyan and Xie, Zhenda and Wei, Y. X. and Wang, Lean and Xiao, Zhiping and Wang, Yuqing and Ruan, Chong and Zhang, Ming and Liang, Wenfeng and Zeng, Wangding},
  booktitle = {Proceedings of the 63rd Annual Meeting of the Association for Computational Linguistics},
  year      = {2025},
  pages     = {23078--23097}
}

@misc{deepseekai2026deepseekv4,
  title  = {{DeepSeek-V4}: Toward Highly Efficient Million-Token Context Intelligence},
  author = {{DeepSeek-AI}},
  year   = {2026},
  note   = {Technical report. Available from \url{https://huggingface.co/deepseek-ai/DeepSeek-V4-Pro}.}
}

@inproceedings{jaegle2021perceiver,
  title     = {Perceiver: General Perception with Iterative Attention},
  author    = {Jaegle, Andrew and Gimeno, Felix and Brock, Andrew and Vinyals, Oriol and Zisserman, Andrew and Carreira, Jo{\~a}o},
  booktitle = {Proceedings of the 38th International Conference on Machine Learning},
  year      = {2021}
}

@inproceedings{alayrac2022flamingo,
  title     = {Flamingo: A Visual Language Model for Few-Shot Learning},
  author    = {Alayrac, Jean-Baptiste and Donahue, Jeff and Luc, Pauline and Miech, Antoine and Barr, Iain and Hasson, Yana and Lenc, Karel and Mensch, Arthur and Millican, Katherine and Reynolds, Malcolm and others},
  booktitle = {Advances in Neural Information Processing Systems},
  year      = {2022}
}

@inproceedings{mu2023gisting,
  title     = {Learning to Compress Prompts with Gist Tokens},
  author    = {Mu, Jesse and Li, Xiang Lisa and Goodman, Noah},
  booktitle = {Advances in Neural Information Processing Systems},
  year      = {2023}
}

@inproceedings{chevalier2023adapting,
  title     = {Adapting Language Models to Compress Contexts},
  author    = {Chevalier, Alexis and Wettig, Alexander and Ajith, Anirudh and Chen, Danqi},
  booktitle = {Proceedings of the 2023 Conference on Empirical Methods in Natural Language Processing},
  year      = {2023}
}

@misc{ge2023incontextautoencoder,
  title         = {In-Context Autoencoder for Context Compression in a Large Language Model},
  author        = {Tao Ge and Jing Hu and Lei Wang and Xun Wang and Si-Qing Chen and Furu Wei},
  year          = {2024},
  eprint        = {2307.06945},
  archivePrefix = {arXiv}
}

@inproceedings{li2021prefix,
  title     = {Prefix-Tuning: Optimizing Continuous Prompts for Generation},
  author    = {Li, Xiang Lisa and Liang, Percy},
  booktitle = {Proceedings of the 59th Annual Meeting of the Association for Computational Linguistics},
  year      = {2021}
}

@inproceedings{rae2019compressive,
  title     = {Compressive Transformers for Long-Range Sequence Modeling},
  author    = {Rae, Jack W. and Potapenko, Anna and Jayakumar, Siddhant M. and Hillier, Chloe and Lillicrap, Timothy P.},
  booktitle = {International Conference on Learning Representations},
  year      = {2020}
}

@misc{makhzani2013ksparse,
  title         = {{k-Sparse} Autoencoders},
  author        = {Alireza Makhzani and Brendan Frey},
  year          = {2014},
  eprint        = {1312.5663},
  archivePrefix = {arXiv}
}

@misc{bricken2023monosemanticity,
  title  = {Toward Monosemanticity: Decomposing Language Models with Dictionary Learning},
  author = {Bricken, Trenton and Templeton, Adly and Batson, Joshua and Chen, Brian and Jermyn, Adam and Conerly, Tom and Turner, Nick and Anil, Cem and Denison, Carson and Askell, Amanda and Lasenby, Robert and Wu, Yifan and Kravec, Shauna and Schiefer, Nicholas and Maxwell, Tim and Joseph, Nicholas and Hatfield-Dodds, Zac and Tamkin, Alex and Nguyen, Karina and McLean, Brayden and Burke, Josiah E. and Hume, Tristan and Carter, Shan and Henighan, Tom and Olah, Christopher},
  year   = {2023},
  note   = {Transformer Circuits Thread, \url{https://transformer-circuits.pub/2023/monosemantic-features/index.html}}
}

@inproceedings{kingma2014auto,
  title     = {Auto-Encoding Variational Bayes},
  author    = {Kingma, Diederik P. and Welling, Max},
  booktitle = {International Conference on Learning Representations},
  year      = {2014}
}

@inproceedings{rezende2014stochastic,
  title     = {Stochastic Backpropagation and Approximate Inference in Deep Generative Models},
  author    = {Rezende, Danilo Jimenez and Mohamed, Shakir and Wierstra, Daan},
  booktitle = {Proceedings of the 31st International Conference on Machine Learning},
  year      = {2014}
}

@misc{qwen2025qwen3,
  title         = {{Qwen3} Technical Report},
  author        = {{Qwen Team}},
  year          = {2025},
  eprint        = {2505.09388},
  archivePrefix = {arXiv}
}

@misc{su2023roformerenhancedtransformerrotary,
  title         = {{RoFormer}: Enhanced Transformer with Rotary Position Embedding},
  author        = {Jianlin Su and Yu Lu and Shengfeng Pan and Ahmed Murtadha and Bo Wen and Yunfeng Liu},
  year          = {2023},
  eprint        = {2104.09864},
  archivePrefix = {arXiv}
}

@misc{zhang2019rootmeansquarelayer,
  title         = {Root Mean Square Layer Normalization},
  author        = {Biao Zhang and Rico Sennrich},
  year          = {2019},
  eprint        = {1910.07467},
  archivePrefix = {arXiv}
}

@inproceedings{dehghani2023scalingvisiontransformers22,
  title     = {Scaling Vision Transformers to 22 Billion Parameters},
  author    = {Dehghani, Mostafa and Djolonga, Josip and Mustafa, Basil and Padlewski, Piotr and Heek, Jonathan and Gilmer, Justin and Steiner, Andreas and Caron, Mathilde and Geirhos, Robert and Alabdulmohsin, Ibrahim and Jenatton, Rodolphe and others},
  booktitle = {Proceedings of the 40th International Conference on Machine Learning},
  year      = {2023}
}

@inproceedings{hsieh2024ruler,
  title         = {{RULER}: What's the Real Context Size of Your Long-Context Language Models?},
  author        = {Hsieh, Cheng-Ping and Sun, Simeng and Kriman, Samuel and Acharya, Shantanu and Rekesh, Dima and Jia, Fei and Zhang, Yang and Ginsburg, Boris},
  booktitle     = {Conference on Language Modeling (COLM)},
  year          = {2024},
  eprint        = {2404.06654},
  archivePrefix = {arXiv}
}

@inproceedings{yen2025helmet,
  title         = {{HELMET}: How to Evaluate Long-Context Language Models Effectively and Thoroughly},
  author        = {Yen, Howard and Gao, Tianyu and Hou, Minmin and Ding, Ke and Fleischer, Daniel and Izsak, Peter and Wasserblat, Moshe and Chen, Danqi},
  booktitle     = {International Conference on Learning Representations (ICLR)},
  year          = {2025},
  eprint        = {2410.02694},
  archivePrefix = {arXiv}
}

@inproceedings{shen2022multilexsum,
  title         = {Multi-{LexSum}: Real-World Summaries of Civil Rights Lawsuits at Multiple Granularities},
  author        = {Shen, Zejiang and Lo, Kyle and Yu, Lauren and Dahlberg, Nathan and Schlanger, Margo and Downey, Doug},
  booktitle     = {Advances in Neural Information Processing Systems Datasets and Benchmarks Track},
  year          = {2022},
  eprint        = {2206.10883},
  archivePrefix = {arXiv}
}

@inproceedings{bai2024longbench,
  title     = {{LongBench}: A Bilingual, Multitask Benchmark for Long Context Understanding},
  author    = {Bai, Yushi and Lv, Xin and Zhang, Jiajie and Lyu, Hongchang and Tang, Jiankai and Huang, Zhidian and Du, Zhengxiao and Liu, Xiao and Zeng, Aohan and Hou, Lei and Dong, Yuxiao and Tang, Jie and Li, Juanzi},
  booktitle = {Proceedings of the 62nd Annual Meeting of the Association for Computational Linguistics (Volume 1: Long Papers)},
  pages     = {3119--3137},
  year      = {2024},
  address   = {Bangkok, Thailand},
  publisher = {Association for Computational Linguistics},
  doi       = {10.18653/v1/2024.acl-long.172},
  url       = {https://aclanthology.org/2024.acl-long.172/}
}

@misc{bai2024longbenchv2,
  title         = {{LongBench v2}: Toward Deeper Understanding and Reasoning on Realistic Long-context Multitasks},
  author        = {Bai, Yushi and Tu, Shangqing and Zhang, Jiajie and Peng, Hao and Wang, Xiaozhi and Lv, Xin and Cao, Shulin and Xu, Jiazheng and Hou, Lei and Dong, Yuxiao and Tang, Jie and Li, Juanzi},
  year          = {2024},
  eprint        = {2412.15204},
  archivePrefix = {arXiv},
  primaryClass  = {cs.CL}
}

@misc{kocetkov2022stack,
  title         = {The {Stack}: 3 {TB} of permissively licensed source code},
  author        = {Kocetkov, Denis and Li, Raymond and Ben Allal, Loubna and Li, Jia and Mou, Chenghao and Mu{\~n}oz Ferrandis, Carlos and Jernite, Yacine and Mitchell, Margaret and Hughes, Sean and Wolf, Thomas and Bahdanau, Dzmitry and von Werra, Leandro and de Vries, Harm},
  year          = {2022},
  eprint        = {2211.15533},
  archivePrefix = {arXiv},
  primaryClass  = {cs.CL}
}

@inproceedings{loshchilov2019adamw,
  title     = {Decoupled Weight Decay Regularization},
  author    = {Loshchilov, Ilya and Hutter, Frank},
  booktitle = {International Conference on Learning Representations (ICLR)},
  year      = {2019}
}

@inproceedings{hu2022lora,
  title     = {{LoRA}: Low-Rank Adaptation of Large Language Models},
  author    = {Hu, Edward J. and Shen, Yelong and Wallis, Phillip and Allen-Zhu, Zeyuan and Li, Yuanzhi and Wang, Shean and Wang, Lu and Chen, Weizhu},
  booktitle = {International Conference on Learning Representations (ICLR)},
  year      = {2022}
}

@inproceedings{zhang2019lookahead,
  title     = {Lookahead Optimizer: $k$ Steps Forward, 1 Step Back},
  author    = {Zhang, Michael R. and Lucas, James and Hinton, Geoffrey and Ba, Jimmy},
  booktitle = {Advances in Neural Information Processing Systems},
  year      = {2019}
}

%%%%%%%%%%%%%%%%%%%%%%%%%%%%%%%%%%%%%%%%%%%%%%%%%%%%%%%%%%%%

\appendix

% =====================================================================
% Appendix sections deferred from Section 3 (Method).
% Drop into the appendix portion of main.tex, after \appendix.
% Labels referenced from the main body:
%   app:architecture, app:pipeline, app:identity-init,
%   app:param-count, app:training, app:baselines.
% The factorization ablation (app:factorization-ablation) and
% architecture ablations (app:arch-ablations) already exist in the
% current draft and are not duplicated here.
% =====================================================================

\tableofcontents

\section{Architecture Details}
\label{app:architecture}

This appendix expands the per-layer compactor description in
\S\ref{sec:method-architecture}. The notation follows the main text:
$d$ is the base model's KV head dimension, $H$ the number of KV-heads
per layer, $d_\ell$ the compactor latent dimension, $t$ the compact
length, $B$ the block count, and $n_c, n_s$ the cross- and
self-attention head counts.

\paragraph{Input concatenation and RoPE handling.}
For a single layer and KV-head, keys and values are concatenated along
the feature axis:
\[
    X^{(h)} = [K^{(h)}; V^{(h)}] \in \mathbb{R}^{T \times 2d},
    \qquad h = 1,\ldots,H.
\]
When RoPE-fix is enabled (the canonical setting), cached keys are first
un-rotated into a position-free frame before this concatenation; values
are not rotated.

\paragraph{Latent queries.}
Each layer maintains learned latents $Z \in \mathbb{R}^{H \times t
\times d_\ell}$. The first axis indexes KV-heads, so each head receives
its own bank of $t$ latents; linear projections are shared across
heads. Sharing the projections gives the compactor an implicit
per-head budget without the parameter blowup that per-head projections
would incur. The canonical initialiser is the block-diagonal
identity-style construction of Appendix~\ref{app:identity-init};
random and orthogonal initialisers are supported as ablations.

\paragraph{Compactor blocks.}
The compactor contains $B$ pre-norm blocks. The standard
Perceiver-resampler setting applies cross-attention only in the first
block; the canonical Still configuration applies cross-attention in
\emph{every} block. Self-attention and the feed-forward sublayer are
independently configurable. With $\mathrm{RMSNorm}$ denoting
non-parametric RMSNorm \citep{zhang2019rootmeansquarelayer}, a block
with all three sublayers computes
\begin{align}
    Z'   &= Z + \mathrm{CrossAttn}(\mathrm{RMSNorm}(Z), X), \\
    Z''  &= Z' + \mathrm{SelfAttn}(\mathrm{RMSNorm}(Z')), \\
    Z''' &= Z'' + \mathrm{FFN}(\mathrm{RMSNorm}(Z'')).
\end{align}
Disabled sublayers omit their residual update; blocks after the first
in the Perceiver-resampler setting apply only the enabled latent
self-attention and feed-forward updates.

\paragraph{Cross-attention.}
Cross-attention is multi-head with $n_c$ heads. We use the convention
that the expanded model dimension is $n_c d_\ell$, so each attention
head has dimension $d_\ell$ rather than $d_\ell / n_c$: multi-head
fan-out widens the projection dimension and the output projection
contracts back to the latent backbone width. The cross-attention
inputs are the RMSNorm'd latent bank
$Q = \mathrm{RMSNorm}(Z) \in \mathbb{R}^{t \times d_\ell}$ (per
KV-head, with projections shared across KV-heads) and the per-head
concatenated cache $X = [K; V] \in \mathbb{R}^{T \times 2d}$. For head
$i$,
\begin{align}
    \tilde{q}_i &= Q W_{Q_i} + b_{Q_i},
        & W_{Q_i} &\in \mathbb{R}^{d_\ell \times d_\ell},
        & b_{Q_i} &\in \mathbb{R}^{d_\ell}, \\
    \tilde{k}_i &= X W_{K_i} + b_{K_i},
        & W_{K_i} &\in \mathbb{R}^{2d \times d_\ell},
        & b_{K_i} &\in \mathbb{R}^{d_\ell}, \\
    v_i &= X W_{V_i},
        & W_{V_i} &\in \mathbb{R}^{2d \times d_\ell}.
\end{align}
The query and key projections carry additive biases; the value
projection does not. Queries and keys are $L_2$-normalized after the
linear projection but before the dot product
\citep[QK-norm;][]{dehghani2023scalingvisiontransformers22}. When
cross-attention positions are supplied, the compactor applies its own
RoPE to the normalized queries and keys:
\begin{align}
    q_i &= \mathrm{RoPE}\!\left(\mathrm{Norm}_{L_2}(\tilde{q}_i),
            \mathbf{p}^{(q)}\right), \\
    k_i &= \mathrm{RoPE}\!\left(\mathrm{Norm}_{L_2}(\tilde{k}_i),
            \mathbf{p}^{(k)}\right).
\end{align}
Key positions $\mathbf{p}^{(k)}$ are the input cache positions; latent
query positions $\mathbf{p}^{(q)}$ are evenly spaced between the first
and last supplied key positions. The compactor RoPE base is a
hyperparameter (canonical value $10$) and can optionally be learned
per dimension.

After QK-norm the raw dot products are cosine similarities. The usual
$1/\sqrt{d_\ell}$ scale produces too-uniform attention in this regime,
so the canonical implementation multiplies logits by $d_\ell$:
\[
    \mathrm{head}_i =
    \mathrm{softmax}\!\left(d_\ell\, q_i k_i^{\top}\right) v_i .
\]
Standard $1/\sqrt{d_\ell}$ and $2 d_\ell$ alternatives are supported as
ablations. Head outputs are concatenated and projected:
\[
    \mathrm{CrossAttn}(Q, X) =
    \mathrm{Concat}(\mathrm{head}_1, \ldots, \mathrm{head}_{n_c}) W_O,
    \qquad W_O \in \mathbb{R}^{n_c d_\ell \times d_\ell}.
\]

\paragraph{Latent self-attention.}
Latent self-attention uses the same expanded-head convention with
$n_s$ heads, the same QK-norm, and the same $d_\ell$ logit scale; no
RoPE is applied:
\[
    \mathrm{SelfAttn}(Z) =
    \mathrm{Concat}(\mathrm{head}_1, \ldots, \mathrm{head}_{n_s})
    W_O^{(\mathrm{sa})}.
\]

\paragraph{Feed-forward sublayer.}
When enabled, the FFN is a two-layer GELU MLP with expansion ratio $r$:
\[
    \mathrm{FFN}(x) = \mathrm{GELU}(x W_1) W_2,
    \qquad
    W_1 \in \mathbb{R}^{d_\ell \times r d_\ell},
    \quad
    W_2 \in \mathbb{R}^{r d_\ell \times d_\ell}.
\]
The canonical configuration omits the FFN.

\paragraph{Output projections.}
Let $Z_\mathrm{out} \in \mathbb{R}^{H \times t \times d_\ell}$ be the
final latent state, optionally after a final RMSNorm. Three independent
linear heads, shared across KV-heads within a layer, produce
\begin{align}
    C_k^{(h)}   &= Z_\mathrm{out}^{(h)} W_\mathrm{key}
                  \in \mathbb{R}^{t \times d}, \\
    C_v^{(h)}   &= Z_\mathrm{out}^{(h)} W_\mathrm{val}
                  \in \mathbb{R}^{t \times d}, \\
    \beta^{(h)} &= Z_\mathrm{out}^{(h)} w_\beta
                  \in \mathbb{R}^{t}.
\end{align}
The only per-KV-head parameters are the latent banks.

\paragraph{Optional $\beta$ shaping.}
Two reparameterizations of $\beta$ are supported but disabled in the
reported Still configuration. \emph{Saturation} clips the bias smoothly:
$\beta \leftarrow \beta_\mathrm{max} \tanh(\beta / \beta_\mathrm{max})$
for $\beta_\mathrm{max} > 0$. \emph{Mass matching} adds the
uniform-attention offset $\log(T/t)$ motivated by
\citet{zweiger2026fast}; in the learnable variant, the implementation
adds $\log(x) + \log(T/t)$ with a positive scalar $x$ initialized at
$10^{-3}$, so this begins as a suppressed mass-matching prior rather
than as exact full mass matching. The paper-facing Still results set
$\beta=0$; the beta-enabled variants in this appendix are diagnostics
of the optional channel.

\paragraph{Per-layer stack and mixed-attention models.}
The full Still module instantiates one compactor object per
transformer layer of the base model. For ordinary global-attention
layers, $(C_k, C_v)$ replace the corresponding full prefix cache in the
reported configuration. For architectures with mixed global and
sliding-window attention, such as Gemma-3, Still applies compaction only
to global-attention layers; sliding-window layers keep their normal
sliding-window cache, preserving the base model's local-attention
semantics while allowing global layers to use the compact prefix.

\paragraph{Canonical configuration summary.}
For reference, the canonical Still configuration on Qwen3-4B used
throughout the paper sets $d_\ell = 256$, $B = 2$ blocks with
cross-attention repeated in every block, $n_c = n_s = 1$ heads, no
FFN, identity-style initialization (Appendix~\ref{app:identity-init})
with RoPE-fix enabled, compactor RoPE base $10$, and the $d_\ell$
logit scale on QK-normed cross-attention. The reported Still results
set the optional bias channel to $\beta=0$; beta-enabled runs in the
appendix are diagnostics. Only $t$ varies across the reported
experiments.

\section{Compaction Pipeline Details}
\label{app:pipeline}

This appendix expands \S\ref{sec:method-pipeline} and
\S\ref{sec:method-iterative}.

\paragraph{Single-pass compaction recipe.}
\begin{enumerate}
    \item Run the frozen base model on the prefix to obtain the full
          per-layer KV cache.
    \item For each layer, select the cache entries to compact (the
          full prefix or, when explicitly requested, a text subrange
          $[s, e)$).
    \item Optionally un-rotate the selected keys with inverse RoPE
          (RoPE-fix).
    \item Concatenate keys and values into $X = [K; V]$ and apply the
          per-layer compactor.
    \item Optionally re-rotate the compact keys at evenly spaced
          output positions over the compacted range.
    \item Splice any preserved prefix or suffix entries back with
          $\beta = 0$.
\end{enumerate}
For paper-facing single-chunk evaluations the training-compatible
layout compacts the full prefill cache used during training and places
the question or continuation on the uncompressed continuation path.
Preserving chat-template tokens separately changes the compact-cache
geometry and is treated as a separate evaluation variant rather than
the default.

\paragraph{Independent chunk compaction.}
The implementation supports splitting a selected text range into
independent chunks, compressing each chunk to $t$ entries, and
concatenating the resulting compact chunks. This is an engineering
baseline only and is not used for any paper-facing iterative result:
each chunk is compressed without seeing compact summaries of earlier
chunks, which is structurally different from the recurrent schedule
below.

\paragraph{Iterative chunked compaction.}
As spelled out in \S\ref{sec:method-iterative}, the natural deployment form of iterative compaction rebuilds the model
cache chunk by chunk: the model prefills a chunk using the current
compact prefix as past key-values, Still compresses the resulting
cache, and the next chunk is processed against that compact prefix.
This matches the cache state seen during streaming inference and
requires repeated LLM forward. This implementation also supports lookahead, a variant that provides a one chunk raw KV cache buffer at prefill. This is the standard procedure used by iterative results in this paper unless stated otherwise.

\paragraph{Bias semantics in attention.}
When the optional bias channel is enabled and the frozen model attends
over a compact prefix followed by any uncompacted suffix, $\beta$ is
added to the compact-prefix logits:
\[
    \mathrm{Attn}\!\left(
        q;\,
        \begin{bmatrix} C_k \\ K_\mathrm{fix} \end{bmatrix},
        \begin{bmatrix} C_v \\ V_\mathrm{fix} \end{bmatrix},
        \begin{bmatrix} \beta \\ \mathbf{0} \end{bmatrix}
    \right)
    =
    \frac{
        \sum_{j=1}^{t}
        \exp\!\left(
            q C_{k,j}^{\top}/\sqrt{d} + \beta_j
        \right) C_{v,j}
        +
        \sum_j
        \exp\!\left(
            q K_{\mathrm{fix}, j}^{\top} / \sqrt{d}
        \right) V_{\mathrm{fix}, j}
    }{
        \sum_{j=1}^{t}
        \exp\!\left(
            q C_{k,j}^{\top}/\sqrt{d} + \beta_j
        \right)
        +
        \sum_j
        \exp\!\left(
            q K_{\mathrm{fix}, j}^{\top} / \sqrt{d}
        \right)
    }.
\]
Non-compact entries carry $\beta = 0$. In the Qwen implementation this
is realized by extending the cache with a per-token bias channel and
adding the bias inside attention for \texttt{CompactedPrefixCache}.
Multi-token prefill uses a FlexAttention score modifier when
available; generation-length queries use an equivalent SDPA fallback
based on dimension augmentation.

\paragraph{Logical vs.\ physical cache length.}
The compact cache separates physical cache length from logical
sequence length. Physically, the prefix has only the compact entries
plus any preserved uncompressed entries. Logically, newly generated or
appended tokens receive RoPE phases as if the full original prefix
were still present. The cache stores an offset equal to the original
sequence length minus the maximum compacted prefix length, and the
model adds this offset to continuation positions before computing
RoPE.

\paragraph{Deployment accounting.}
Still is a retained-cache reduction method in the implementation
reported here. For single-pass compaction, the base model first
prefills the original prefix and materializes the full KV cache; Still
then maps that cache to a shorter \texttt{CompactedPrefixCache}. A
serving implementation can release the original prefix cache as soon as
the compact cache has been constructed, although PyTorch may keep the
freed blocks in its CUDA allocator for reuse. The peak live tensor
state during this transition is therefore base-model weights plus the
full prefix KV cache, the compact KV cache, compactor weights, and
transient compactor activations. The retained serving state after the
transition is only the compact physical cache plus any generated suffix
tokens.

For a transformer with \(L\) layers, \(H_\mathrm{kv}\) KV heads, head
dimension \(d\), element size \(b\) bytes, and physical cache length
\(S\), the KV cache occupies
\[
    M_\mathrm{KV}(S) = 2 L H_\mathrm{kv} d b S
\]
bytes, counting keys and values. For Qwen3-4B with
\(L=36\), \(H_\mathrm{kv}=8\), \(d=128\), and bf16/fp16 KV tensors,
this is about \(144\) KiB per token: \(1.125\) GiB at \(8\)k tokens,
\(4.5\) GiB at \(32\)k, \(9.0\) GiB at \(64\)k, and \(18.0\) GiB at
\(128\)k. The same formula with \(S=K\) gives the compact-cache
footprint; for example, \(K=328\) stores about \(46\) MiB of KV state.
The canonical Qwen3-4B compactor at \(t=128\) adds roughly \(50\)M
parameters, about \(1\%\) of the base model and about \(95\) MiB in
bf16.

This accounting is about retained memory rather than production serving
measurements. Because subsequent tokens attend to the physical cache
length \(K\), not the logical prefix length \(T\), Still reduces the
retained KV state and the attention operand size after compaction.
The single-pass path still includes the full prefill before the
compactor forward, so the result should be read as cache-state
reduction rather than elimination of the initial full-context pass.
Finally, the reported Still results set the optional bias channel to
\(\beta=0\). Enabling additive
\(\beta\) in production requires attention-kernel support or a custom
attention path; the current repo implements it in a patched
HF/PyTorch path, not as an unmodified vLLM or FlashAttention feature.

\subsection{vLLM cache-injection microbenchmark}
\label{app:vllm-injection}

To check that the accounting above corresponds to an executable serving path,
we implemented a vLLM~0.11.0 prototype for Qwen3-4B on a single NVIDIA H200. The
prototype includes a standard full-cache vLLM row and compact-cache rows that
allocate only \(K\) physical prefix slots in vLLM's paged KV cache, copy compact
K/V tensors into those slots, and decode the continuation in vLLM. During
compact-cache decode, the physical cache length is \(K\), while the RoPE
positions are offset by \(T-K\), matching the logical prefix semantics described
above. The Still row injects Still's synthesized \(C_k,C_v\); the KV-Distill row
injects the selected K/V rows produced by the corrected KV-Distill
LoRA/scorer checkpoint. Both compact rows intentionally omit \(\beta\): they
test concrete vLLM K/V injection, not the optional additive-bias diagnostic.

\begin{table}[h]
\centering
\caption{Concrete vLLM timings for full and compact cache states on Qwen3-4B at
64k context. Each row uses one H200, bf16 K/V tensors, one request, and exactly
128 generated tokens. TTFT includes all work through the first generated token;
for compact rows this includes full-prefix prefill, compaction or selection,
K/V installation into vLLM, and the first vLLM decode step. Decode throughput is
post-first-token throughput. Peak allocation is the measured CUDA allocation
inside this vLLM prototype, including model and cache state.}
\label{tab:vllm-still-injection}
\scriptsize
\begin{tabular}{lrrrrr}
\toprule
Condition & Retained KV & Peak alloc. & TTFT s & E2E s & Decode tok/s \\
\midrule
Full cache & 9.000 GiB & 121.4 GiB & 3.147 & 4.372 & 103.6 \\
Still compact & 0.180 GiB & 57.4 GiB & 6.757 & 8.078 & 96.1 \\
KV-Distill compact & 0.180 GiB & 57.4 GiB & 11.361 & 12.608 & 101.9 \\
\bottomrule
\end{tabular}
\end{table}

At this single-request setting, compact-cache serving is not an end-to-end
latency win because the prototype still pays the full prefill and an
out-of-band cache construction and copy step; under these measured single-turn
conditions there is therefore no latency break-even point to report. The
retained serving state is the intended advantage: both compact rows retain
\(0.180\) GiB of K/V state, compared with \(9.0\) GiB for the full 64k cache,
and reduce the measured prototype peak allocation from \(121.4\) GiB to
\(57.4\) GiB while decoding from injected compact caches at roughly the same
post-first-token throughput as full-cache vLLM. A production batching,
concurrency, or sparse-kernel comparison would require a scheduler-level
integration beyond this cache-injection microbenchmark.

\section{Identity-Style Initialization}
\label{app:identity-init}

Random initialization makes the untrained compactor far from a
pass-through cache, which makes it difficult to isolate bugs in RoPE
handling, optional $\beta$ injection, and cache integration during early
training. Still therefore supports an identity-style initialization
that makes the compactor a near-pass-through map at $t = T$. This is a
diagnostic and training-stabilization device, not an exact algebraic
identity for all inputs.

\paragraph{Construction.}
The construction requires $d_\ell = 2d$. In the single-cross-head case,
the value projection and cross-attention output projection are
initialized as identity maps over the concatenated $[K; V]$
representation. The output heads extract the key and value halves:
\[
    W_\mathrm{key} = [I_d \; 0],
    \qquad
    W_\mathrm{val} = [0 \; I_d],
    \qquad
    w_\beta = 0.
\]
For multiple cross-attention heads, this identity path is placed in
the first expanded head; the remaining expanded heads are initialized
so that they do not contribute at step zero.

\paragraph{Query/key pathway.}
The query/key pathway is initialized so that cross-attention is
dominated by relative position rather than content. With biased
identity initialization the latent bank is zeroed, the cross-attention
key projection is zeroed, and query and key biases are aligned to a
fixed unit direction. After QK-norm, the unrotated query and key
directions are therefore nearly constant, and the compactor's internal
RoPE is the main signal distinguishing positions. With a small RoPE
base ($10$ in the canonical configuration), this produces a sharply
local, block-diagonal attention pattern when the latent positions tile
the input range. At $t < T$, each latent reads from a local region
rather than a single token.

\paragraph{Refinement sublayers as residual identities.}
To avoid disturbing the initial pass-through path, the self-attention
output projections, FFN second layers, and any later cross-attention
output projections are zero-initialized. These refinement sublayers
therefore begin as residual identities and become active only as
their output projections train away from zero.

\section{Parameter Count}
\label{app:param-count}

Let $L$, $H$, $d$ be the base model's layer count, KV-head count, and
KV head dimension; let the compactor have latent dimension $d_\ell$,
compact length $t$, $B$ blocks, $n_c$ cross-attention heads, $n_s$
self-attention heads, and FFN expansion ratio $r$. Let $B_c$ be the
number of blocks containing cross-attention: $B_c = 1$ for the
Perceiver-resampler setting and $B_c = B$ when cross-attention is
repeated every block. Ignoring small optional biases, the per-layer
parameter count is
\[
\begin{aligned}
    p_\ell
    &= \underbrace{H t d_\ell}_{\text{latent bank}}
    + B_c \underbrace{\left(
        2 n_c d_\ell^2 + 4 d n_c d_\ell
    \right)}_{\text{cross-attention}} \\
    &\quad
    + B \underbrace{\left( 4 n_s d_\ell^2 \right)}_{\text{self-attention}}
    + B \underbrace{\left( 2 r d_\ell^2 \right)}_{\text{FFN, if enabled}}
    + \underbrace{\left( 2 d_\ell d + d_\ell \right)}_{\text{output heads}}.
\end{aligned}
\]
Disabled sublayers drop their term. The total parameter count is
$L p_\ell$.

For Qwen3-4B with $L = 36$, $H = 8$, $d = 128$, the canonical
configuration $(t = 128, d_\ell = 256, B = 2, n_c = n_s = 1,$
repeated cross-attention, no FFN$)$ has approximately $50$M
parameters, around $1\%$ of the base model. Larger illustrative
configurations include $(t = 140, d_\ell = 256, B = 4, n_c = n_s = 8,
r = 4)$ at approximately $466$M parameters with first-block-only
cross-attention or $692$M with cross-attention repeated in every
block. These counts include the expanded multi-head projections used
by the implementation.

\section{Training Details}
\label{app:training}

\paragraph{Training corpus.}
The canonical training corpus is a four-domain extractive MCQ dataset
covering Financial filings, Project Gutenberg literature, Legal
documents, and Code. Each item is structured as
\[
\texttt{(ctx\_ids, prompt\_ids, answer\_ids)},
\]
with \texttt{ctx\_ids} the long-context prefill, \texttt{prompt\_ids}
the MCQ question, and \texttt{answer\_ids} the answer-letter target.
Questions are constructed from random $1$k-token sub-chunks of the
prefill, verified by the same frozen base model used as teacher, and
filtered to remove items the base model can answer without the
context.

For the canonical $8$k-context Qwen3-4B run, raw per-domain item
counts are:
\begin{center}
\begin{tabular}{lr}
\toprule
Domain & Items \\
\midrule
Financial & $236{,}808$ \\
Project Gutenberg & $110{,}567$ \\
Code & $38{,}394$ \\
Legal & $30{,}136$ \\
\midrule
Capped per-domain (training) & $30{,}136$ \\
Total training items & $120{,}544$ \\
\bottomrule
\end{tabular}
\end{center}
Domains are downsampled to the smallest, giving roughly $1.0$B
context tokens at $8$k context length before prompt and answer tokens.

\paragraph{Loss details.}
The training signal is forward KL from the full-context teacher to
the compact-cache student, evaluated only on answer-side tokens. The
KL is computed against the top-$200$ teacher-vocabulary tokens with
the gold answer token forced into the support; this keeps the per-step
memory cost bounded. Appendix~\ref{app:kl-support-ablation} reports a
small retraining ablation showing that top-$200$ is not worse than
top-$1000$ or full-vocabulary support under a common full-vocabulary
evaluation. Prompt and pad tokens are masked out of the loss.

%We should indicate context length and num latents here
\paragraph{Optimization.}
\begin{center}
\begin{tabular}{ll}
\toprule
Optimizer & AdamW, $\beta_1 = 0.9$, $\beta_2 = 0.95$ \\
Learning rate & $4 \times 10^{-5}$ \\
Warmup & $100$ steps, linear from $10^{-6}$ \\
Schedule & Constant after warmup \\
Weight decay & $0.01$ on matrix params, $0$ on biases / 1D params \\
Gradient clipping & $1.0$ \\
Microbatch & $4$ per GPU \\
GPUs & $8 \times$ NVIDIA H200 (DDP) \\
Gradient accumulation & $1$ \\
Effective batch & $32$ \\
Headline checkpoint & step $1500$ \\
\bottomrule
\end{tabular}
\end{center}
The compactor parameters are the only trainable parameters; base-model
weights are frozen throughout. Except for the explicitly labeled
generation-aligned LongBench checkpoint in \S\ref{subsec:generation},
paper-facing Qwen3-4B Still numbers use the step-$1500$ checkpoint. The above refers to 256 latents on 32k sequence length.

\paragraph{Compute accounting.}
All reported training and evaluation jobs were run on managed cloud GPU workers. The main Qwen3-4B Still and KV-Distill checkpoints
use $8 \times$ NVIDIA H200 workers (141GB HBM per GPU). Architecture,
factorization, and latent-count ablations use the same step-$1500$
cutoff and the same class of workers; some replacement runs use
$4 \times$ H200 workers with adjusted microbatching to keep the
effective batch at $32$. Scale-transfer runs use H200 workers with
model-specific microbatch and gradient-checkpointing settings. The
Still, KV-Distill, H2O, SnapKV, StreamingLLM, HELMET, RULER,
LongBench, and summarization evaluations use one to four H100/H200
workers depending on model size and context length. The paper-facing
numbers exclude preliminary failed or diagnostic jobs, which required
additional exploratory compute beyond this reproduction budget.

\paragraph{Iterative Compaction}
To implement iteration, we train over multiple passes using the same compactor at each pass. We found that a reduced learning rate of 2e-4 was necessary in this setting. The Lookahead optimizer \citep{zhang2019lookahead} improves training stability. Alternatively, competitive results can be achieved without Lookahead by adding an auxiliary loss at every prefill.

For the aux-KV iterative variant,
we optimize a joint objective:
\begin{equation}
\mathcal{L}
=
\mathcal{L}_{\text{MCQ}}
+
\lambda_{\text{aux}} \, \mathcal{L}_{\text{aux-kv}},
\qquad
\lambda_{\text{aux}} = 0.5.
\end{equation}
The auxiliary term is computed on all passes except the first, by matching
student chunk K/V against teacher full-prefill chunk K/V.
For each pass $p$ and layer $\ell$, we compute normalized relative errors
\begin{equation}
r^{(K)}_{p,\ell}
= \sqrt{\frac{\operatorname{mean}\!\left((K_s - K_t)^2\right)}
             {\max\!\left(\operatorname{mean}(K_t^2),\, \epsilon\right)}},
\qquad
r^{(V)}_{p,\ell}
= \sqrt{\frac{\operatorname{mean}\!\left((V_s - V_t)^2\right)}
             {\max\!\left(\operatorname{mean}(V_t^2),\, \epsilon\right)}},
\qquad
\epsilon = 1e-8
\end{equation}
and apply a Huber loss averaged over layers and passes:
\begin{equation}
\mathcal{L}_{\text{aux-kv}}
=
\sum_{p=1}^{N} \tilde{w}_p \,
\frac{1}{L} \sum_{\ell=1}^{L}
\frac{
  \operatorname{Huber}_\delta\!\left(r^{(K)}_{p,\ell}\right)
  +
  \operatorname{Huber}_\delta\!\left(r^{(V)}_{p,\ell}\right)
}{2},
\qquad \delta = 1.0,
\end{equation}
where $w_p \propto (N - p + 1)$, normalized as $\tilde{w}_p = w_p / \sum_j w_j$,
so that earlier passes receive larger weight. Empirically, this variant remains competitive through 32k context but does not match the quality of its Lookahead counterpart (Figure~\ref{fig:iter-quality-appendix}).

\section{Top-$k$ KL Support Ablation}
\label{app:kl-support-ablation}

We check whether the top-$200$ teacher-vocabulary support used in the
main training loss is an overly aggressive approximation. We retrain
four Qwen3-4B Still compactors on the same four-domain MCQ mixture at
$4$k context length, $512$ latents, and $1000$ steps, varying only the
teacher KL support: top-$50$, top-$200$, top-$1000$, or the full
vocabulary. We then evaluate all final checkpoints on the same $200$
held-out MCQ examples using full-vocabulary KL and compact-cache
cross-entropy.

\begin{table}[h]
\centering
\scriptsize
\caption{Small retraining ablation for the MCQ KL vocabulary support. Under a
common full-vocabulary evaluator, top-$200$ is not worse than top-$1000$ or
full-vocabulary training support.}
\label{tab:kl-support-ablation}
\begin{tabular}{lccc}
\toprule
Training KL support & Full-vocab KL $\downarrow$ & Compact CE $\downarrow$ & MCQ acc. (\%) $\uparrow$ \\
\midrule
Top-$50$ & $0.3834$ & $0.4953$ & $92.5$ \\
Top-$200$ & $\mathbf{0.3260}$ & $\mathbf{0.4352}$ & $\mathbf{94.0}$ \\
Top-$1000$ & $0.3407$ & $0.4520$ & $\mathbf{94.0}$ \\
Full vocab & $0.3424$ & $0.4534$ & $90.0$ \\
\bottomrule
\end{tabular}

\end{table}

\section{Baseline Implementation Details}
\label{app:baselines}

This appendix records the exact configuration used for each baseline
in \S\ref{sec:method-baselines}.

\paragraph{H2O \citep{zhang2023h2o}.}
Reference queries are extracted via repeat-prefill on the same prefix,
matching the protocol of \citet{zweiger2026fast}. Token retention
scores are cumulative attention weights from the extracted queries;
the top-$K$ tokens by cumulative score are kept. Values are taken
directly from the retained source positions (\texttt{c2\_method=direct})
and no $\beta$ correction is applied. The cache budget $K$ is varied
across experiments to match Still's compression ratio.

\paragraph{SnapKV \citep{li2024snapkv}.}
Token scoring restricts attention to a $64$-token observation window
of recent positions, with a $5$-token pooling kernel applied to the
score vector before top-$K$ selection. Retained keys and values are
taken directly from the source positions; no $\beta$ correction.

\paragraph{StreamingLLM \citep{xiao2024efficient}.}
The cache consists of $4$ attention-sink tokens at the start of the
prefix together with the most recent $K - 4$ tokens, for a total
budget of $K$. Sinks are taken from the absolute first positions of
the prefix in the un-rotated cache, matching the construction of
\citet{xiao2024efficient}. No $\beta$ correction.

\paragraph{Attention Matching \citep{zweiger2026fast}.}
Reference queries are extracted by re-prefilling the original context
through the frozen base model, following \citet{zweiger2026fast}. The
primary Attention Matching curve uses the highest-attention-keys
variant: key selection takes the top-$K$ source positions by RMS
attention score, compact values are reconstructed by least squares
against the extracted queries, and the per-token bias $\beta$ is
fitted by bounded nonnegative least squares with two refinement
iterations. We also implemented the paper's Orthogonal Matching
Pursuit family, including the batched OMP schedule. The dashed
AM-OMP-fast curve in Figure~\ref{fig:pareto-frontier} is this
diagnostic variant: it is useful for checking the high-quality side of
the per-context-synthesis frontier, but it is not uniformly more accurate
than the top-$K$ variant on our Qwen3-4B sweep. The Still--AM cell-level comparison in
Appendix~\ref{app:still-vs-am-delta} therefore uses the measured
top-$K$ Attention Matching curve.

\subsection{KV-Distill fairness and optimization}
\label{app:kvdistill-fairness}

KV-Distill is trained on Qwen3-4B with the same four-domain
corpus and matching context-length splits as Still; we do not use
released checkpoints. A learned scorer reads frozen layer-$6$ hidden
states and selects the top-$K$ tokens, with the first $8$ tokens forced
as sinks. Rank-$128$ LoRA adaptors are attached to $Q,K,V,O$; $Q/O$
adaptors are applied only to selected tokens and $K/V$ adaptors to the
full prefill. The adapted model is run over the full context, the
selected K/V rows are sliced from that adapted cache, and the frozen
base model then answers against the compressed cache. Training uses
symmetric KL distillation ($\lambda = 0.6$) on answer-side tokens plus
an auxiliary scorer loss toward key-RMS selection targets. The
question/answer path is left uncompressed at inference, matching
Still's evaluation path. The paper-facing checkpoints are taken at
step $1500$ for the matched sweeps rather than selected per cell by
test accuracy. We note that the loss plateaus for both before 1500 for all configurations.

\begin{table}[h]
\centering
\scriptsize
\caption{Audit of the matched Still--KV-Distill comparison.}
\label{tab:kvdistill-audit}
\begin{tabular}{@{}p{0.22\linewidth}p{0.35\linewidth}p{0.35\linewidth}@{}}
\toprule
Axis & Still & KV-Distill \\
\midrule
Training data & Same Qwen3-4B four-domain corpus for the mixed-domain sweeps; RULER-trained results use the same RULER train split as KV-Distill. & Same matching corpus and context-length split as the paired Still run; no released checkpoints are used. \\
Training signal & Forward KL from the full-context teacher to the compact-cache student on answer-side tokens. & Symmetric KL on answer-side tokens plus a scorer BCE auxiliary loss toward key-RMS token targets. \\
Trainable parameters & Per-layer compactor only; the base model remains frozen. The canonical Qwen3-4B compactor is roughly $50$M parameters at $t=128$ and scales with compact length. & Rank-$128$ LoRA adapters on $Q,K,V,O$ plus a learned token scorer. Base weights remain frozen, but LoRA adapters are persistent parameter overhead and count as model modification. \\
Prefill and decoding path & Full prefix is prefetched by the frozen base model, then replaced by the compact $(C_k,C_v)$ cache; reported Still results use $\beta=0$. & The selected positions are chosen by the scorer, an adapted full-context prefill is run with LoRA enabled and selected-token masking, selected adapted K/V rows are sliced, and decoding runs with LoRA disabled. \\
Continuation path & Question/answer tokens are left uncompressed. & Same uncompressed question/answer path as Still. \\
Budget policy & Figures match retained slots $K$; no extra slots are added for parameter overhead. & Figures match the same retained slots $K$. We do not give KV-Distill a larger retained budget to offset Still's compactor parameters, and reported Still has no beta-cache overhead. \\
\bottomrule
\end{tabular}
\end{table}

\subsection{Baseline selection and tuning policy}
The plotted matrix keeps one configuration per method across tasks and
context lengths, with only the cache budget $K$ varying to match the
target compression ratio. This fixed policy avoids per-cell tuning on
the evaluation grid, but it is not a default-only implementation pass:
we tried the main baseline-strengthening choices in
Table~\ref{tab:baseline-tuning} and then fixed one paper-facing policy
per method.

\begin{table}[h]
\centering
\scriptsize
\caption{Baseline tuning and final fixed policies.}
\label{tab:baseline-tuning}
\begin{tabular}{@{}p{0.16\linewidth}p{0.40\linewidth}p{0.34\linewidth}@{}}
\toprule
Method & Tuning surface tried & Final paper-facing policy \\
\midrule
H2O & Multiple observation/reference-query settings, including local-window variants and repeat-prefill reference queries. & Repeat-prefill reference queries, cumulative-attention top-$K$ selection, direct retained values, no $\beta$, and a no-local-window prefill-then-compact path. \\
SnapKV & Multiple observation-window and pooling-kernel settings around the published SnapKV scoring rule. & Published-style $64$-token observation window, $5$-token pooling kernel, direct retained values, no $\beta$, and a no-local-window outer evaluation path. \\
StreamingLLM & Multiple sink-count settings for the sink-plus-recent-token policy. & $4$ absolute prefix sink tokens plus the most recent $K-4$ tokens, direct retained values, and no $\beta$. \\
Attention Matching & Highest-attention-keys selection and the Orthogonal Matching Pursuit family, including the batched OMP schedule, with reference-query value fitting and bounded $\beta$ fitting. & Top-$K$ RMS highest-attention-keys selection with least-squares values and bounded NNLS $\beta$ fitting; OMP/OMP-fast is reported as a diagnostic curve because it was not uniformly more accurate. \\
KV-Distill & LoRA rank, scorer layer, forced-sink count, training-step choice, and the routed-cache training/evaluation path. & Rank-$128$ Q/K/V/O LoRA, layer-$6$ scorer, $8$ forced sinks, selected-cache slicing from an adapted prefill, and matched step-$1500$ checkpoints rather than per-cell checkpoint selection. \\
\bottomrule
\end{tabular}
\end{table}

For heuristic selection baselines, the final policies use the published
operating points that define the methods in this setting together with
the shared evaluation path. For Attention Matching and KV-Distill, we
include the strongest matched variants we implemented under the same
Qwen3-4B model family, data mixture, cache budget grid, and uncompressed
question/answer continuation path used by Still. The fixed checkpoint
policy is also deliberate: paper-facing KV-Distill cells use the
matched step-$1500$ checkpoint for each run, not the best checkpoint
selected by test or validation accuracy.

\paragraph{Additional baselines considered.}
PyramidKV appears in the design-space taxonomy because it is an
important selection baseline, but it is not included in the reported
Qwen3-4B numeric matrix. KVZip \citep{kim2025kvzip} is closer to
Still's deployment regime because it scores cache entries without
knowing the future user query. We therefore add a targeted 8k/16k
KVZip comparison under the same Qwen3-4B MCQ protocol as the other
frontier baselines; Appendix~\ref{app:kvzip-partial} gives the
implementation details, partial results, and scope rationale. Two
recent query-agnostic or learned-eviction methods remain especially
relevant for future matched comparisons: Compactor
\citep{chari2024compactor}, which uses approximate leverage scores
with context calibration and a vLLM-oriented sparse-cache
implementation, and KVP \citep{moschella2026learning}, which trains
lightweight per-head policies to rank tokens for future utility. We
do not claim numerical dominance over baselines not run under the
shared Qwen3-4B protocol above.

\section{Latent Capacity and Context-Length Sweeps}
\label{app:capacity-sweeps}
 
The compactor exposes two scaling knobs at fixed architecture: the latent count $t$, which controls compression ratio at a given input length, and the input length $T$ itself. The single-sentence claim in \S\ref{subsec:design-space} (compact-cache utility rises monotonically with latent count and is mostly stable at fixed compression across context lengths) compresses three sweeps, which we report in detail below. All sweeps use the canonical Qwen3-4B configuration of \S\ref{sec:method-architecture} and the multidomain training corpus of \S\ref{app:training} unless otherwise stated; tail-mean conventions follow the figure captions.
 
\subsection{Latent count at fixed context length}
\label{app:latent-sweep}

\begin{figure}[h]
    \centering
    \includegraphics[width=\linewidth]{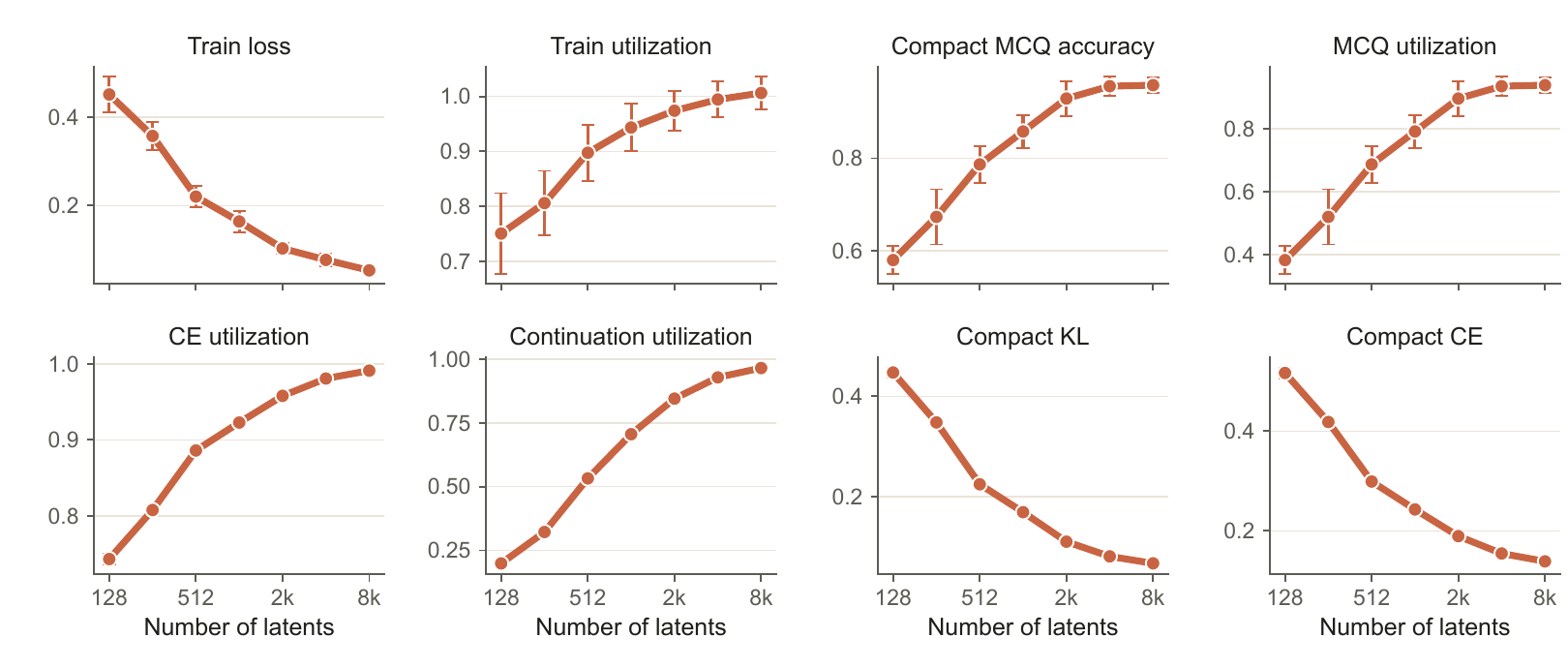}
    \caption{Single-domain latent sweep. Eight metrics as a function of the
    number of latents $t$ (log$_2$ axis), holding training context length
    fixed at $8192$ tokens. Top row: train-loss and train-utilization
    tail-means over the final $10\%$ of training steps. Bottom row: held-out
    compact MCQ accuracy, MCQ utilization, CE utilization, continuation
    utilization, compact KL and compact CE, each as tail-means over the
    final $30\%$ of evaluation steps. Compact MCQ accuracy and CE
    utilization track each other smoothly across $128$--$2048$ latents.
    Bands give one tail standard deviation.}
    \label{fig:latent-sweep-single-domain}
\end{figure}
 
We sweep latent counts $t \in \{32, 64, 128, 256, 512, 1024, 2048\}$ at fixed context length $T = 8192$, training a fresh compactor at each setting under the canonical recipe. Figure~\ref{fig:latent-sweep-single-domain} reports eight metrics on the single-domain MCQ benchmark. Compact MCQ accuracy rises monotonically with $t$, with the steepest gains between $128$ and $1024$ latents and diminishing returns thereafter --- the gap between $t = 1024$ and $t = 2048$ is small relative to the gap between $t = 256$ and $t = 1024$. CE utilization tracks compact MCQ accuracy across the entire sweep. This is the basic empirical condition under which the proxy loss Still actually optimizes during training (forward KL on answer-side tokens) and the downstream extractive metric (multiple-choice answer accuracy) move together: optimizing for the proxy is faithful to the deployment objective at every operating point in this regime.
 
\begin{figure}[h]
    \centering
    \includegraphics[width=\linewidth]{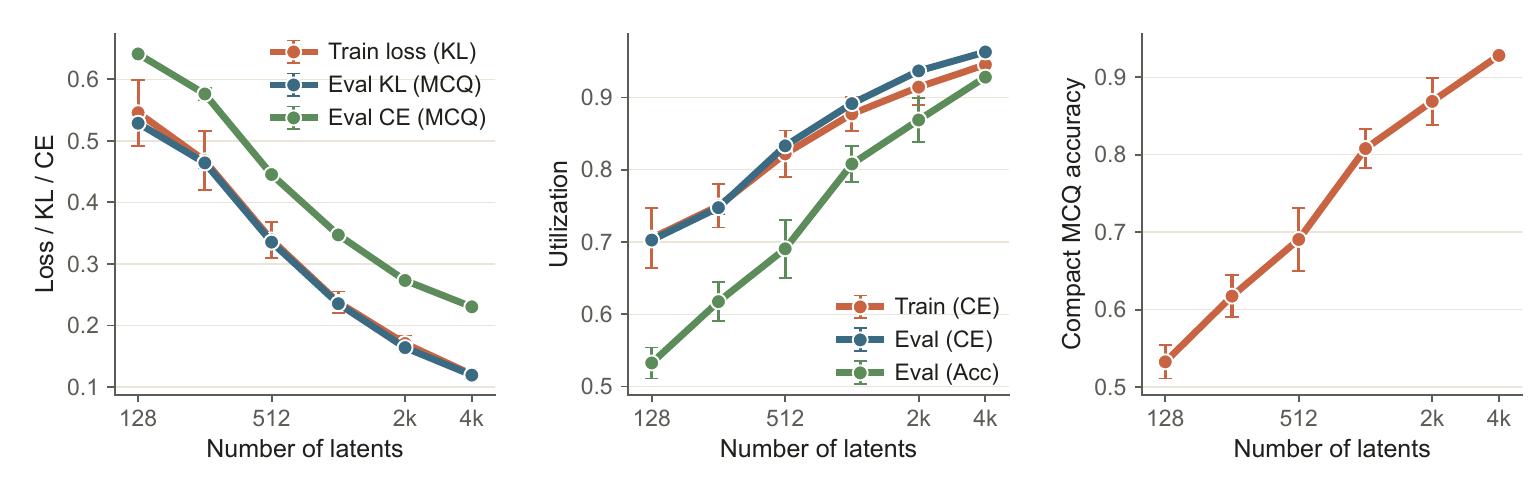}
    \caption{Multidomain latent sweep --- aggregate metrics. Train loss, eval KL, and eval CE (left); utilization under three definitions (center); and compact MCQ accuracy (right) as a function of latent count $t$, on the multidomain benchmark (Financial, Legal, Project Gutenberg, Code) at $8192$-token context. The plotted eval aggregate is reconstructed from the per-domain columns available for this figure (Financial, Gutenberg, and Legal); Code is part of the same multidomain training and evaluation corpus. Compact MCQ accuracy rises monotonically with $t$ and is within $\sim 4$ points of full context at $t = 1024$ ($8\times$ compression).}
    \label{fig:memopt-metrics}
\end{figure}
 
The same sweep on the multidomain corpus (Figure~\ref{fig:memopt-metrics}) shows the same shape: train loss, eval KL, and eval CE all decrease smoothly with $t$; the three utilization definitions all rise toward the full-context ceiling; and compact MCQ accuracy reaches within $\sim 4$ points of the full-context baseline at $t = 1024$ --- the canonical $8\times$ compression operating point. The multidomain numbers are slightly lower than single-domain at matched $t$, consistent with the multidomain corpus being harder on average, but the qualitative scaling is identical.
 
\begin{figure}[h]
    \centering
    \includegraphics[width=0.6\linewidth]{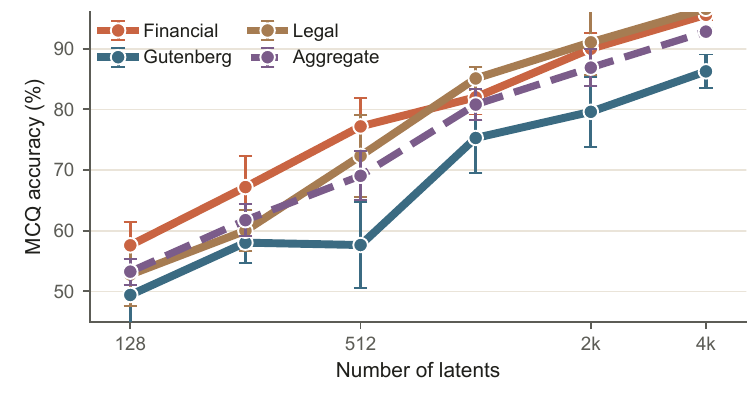}
    \caption{Per-domain MCQ accuracy across the latent sweep. Compact MCQ accuracy on the displayed Financial, Gutenberg, and Legal slices as a function of latent count, with a dashed line for their plotted aggregate. The underlying multidomain sweep also includes Code.}
    \label{fig:memopt-per-domain}
\end{figure}
 
The displayed per-domain breakdown (Figure~\ref{fig:memopt-per-domain}) confirms that the aggregate scaling is not a domain-mixing artefact. Financial, Gutenberg, and Legal each scale smoothly with $t$ and converge above $80\%$ compact accuracy by $t = 1024$. Code is included in the multidomain corpus used for this sweep; the plotted diagnostic focuses on the three displayed domains for readability rather than changing the training or evaluation mixture.
 
\subsection{Variable-context evaluation at fixed checkpoint}
\label{app:variable-context}
 
\begin{figure}[h]
    \centering
    \includegraphics[width=0.6\linewidth]{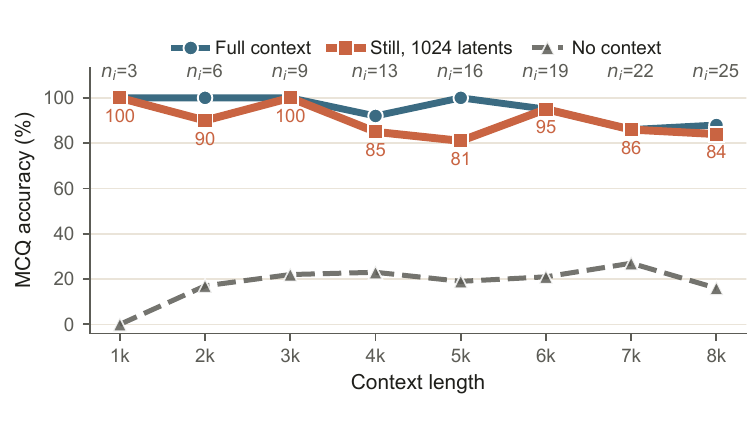}
    \caption{A single $1024$-latent checkpoint generalizes across context lengths it was not trained on. The canonical multidomain $1024$-latent compactor (trained at $T = 8192$) evaluated on held-out MCQs bucketed by document length from $1$k to $8$k tokens.}
    \label{fig:variable-context-eval}
\end{figure}
 
A potential concern with the sweep above is that each $(T, t)$ point is a separately trained compactor, and the smooth scaling we observe could be a property of the training distribution rather than of the compactor itself --- a checkpoint trained at one context length might collapse on shorter or longer inputs. Figure~\ref{fig:variable-context-eval} addresses this directly. We take a single canonical checkpoint --- the multidomain $1024$-latent compactor trained at $T = 8192$ --- and evaluate it on held-out MCQs bucketed by document length from $1$k to $8$k tokens, with no retraining or per-bucket adaptation. Compact accuracy holds up across the full bucket range. This is the basic generalization condition for using a single checkpoint at deployment time, where input lengths vary continuously and per-input retraining is not an option.

\begin{figure}[h]
    \centering
    \includegraphics[width=0.82\linewidth]{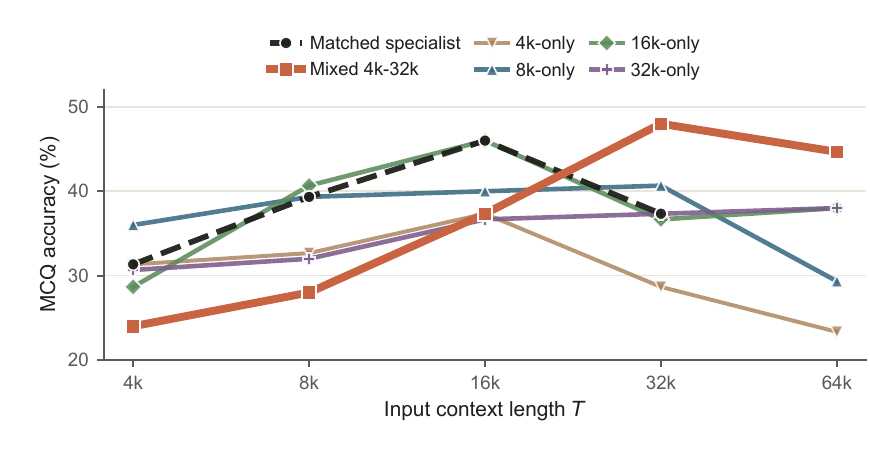}
    \caption{Fixed-cache variable-input-length study at $t=512$. We train specialists at $4$k, $8$k, $16$k, and $32$k, plus a single mixed-length checkpoint over the same range, then evaluate all checkpoints from $4$k to $64$k. The dashed matched-specialist reference uses the checkpoint trained at the corresponding length and is undefined at the extrapolated $64$k point.}
    \label{fig:variable-t-input-lengths}
\end{figure}

We then run a stricter fixed-cache study in which the compact budget is held at $t=512$ while input length varies over $T \in \{4\mathrm{k}, 8\mathrm{k}, 16\mathrm{k}, 32\mathrm{k}\}$ during training and up to $64\mathrm{k}$ at evaluation. Figure~\ref{fig:variable-t-input-lengths} compares per-length specialists to single-length checkpoints and to one mixed-length checkpoint trained with scaled compactor positions and a length/compression embedding. The mixed checkpoint pays a short-context tax relative to specialists at $4$k and $8$k, but is strongest at the long end, reaching $48.0\%$ at $32$k and $44.7\%$ at the held-out $64$k point. By contrast, the $8$k-only checkpoint reaches $40.7\%$ at $32$k but falls to $29.3\%$ at $64$k. Thus mixed-$T$ training gives a useful one-checkpoint deployment point for long inputs, but it does not yet remove the quality cost of length specialization at short contexts.

\subsection{Context length at fixed compression ratio}
\label{app:context-sweep}
 
\begin{figure}[h]
    \centering
    \includegraphics[width=\linewidth]{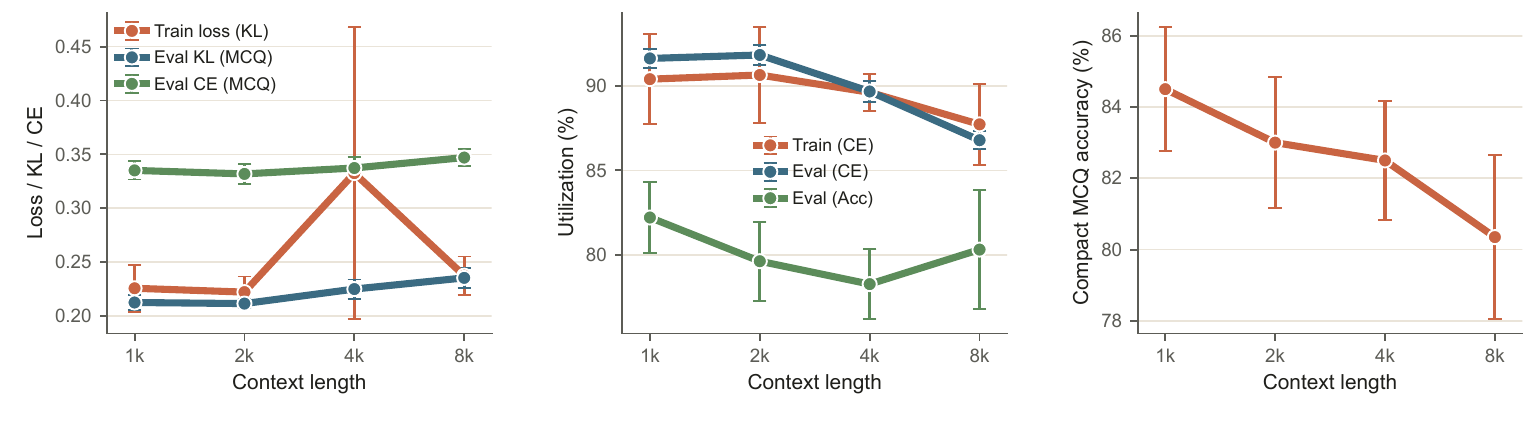}
    \caption{Context-length sweep at fixed $8\times$ compression. Aggregate metrics as we scale the training context length $T$ from $1$k to $8$k while holding the compression ratio $T/t = 8$ fixed (i.e.\ $128, 256, 512, 1024$ latents respectively).}
    \label{fig:context-sweep-metrics}
\end{figure}
 
\begin{figure}[h]
    \centering
    \includegraphics[width=0.55\linewidth]{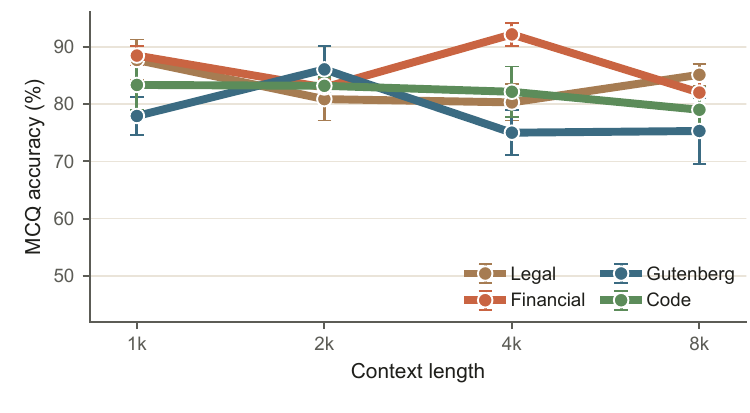}
    \caption{Per-domain accuracy at fixed $8\times$ compression across context lengths. Compact MCQ accuracy on each of the four multidomain domains as we scale $(T, t) \in \{(1\text{k}, 128), (2\text{k}, 256), (4\text{k}, 512), (8\text{k}, 1024)\}$.}
    \label{fig:context-sweep-per-domain}
\end{figure}
 
The complementary question is what happens when we hold compression ratio fixed and vary absolute context length. We sweep $(T, t) \in \{(1\text{k}, 128), (2\text{k}, 256), (4\text{k}, 512), (8\text{k}, 1024)\}$, training a separate compactor at each setting under the canonical recipe. Figures~\ref{fig:context-sweep-metrics} and~\ref{fig:context-sweep-per-domain} report the result.
 
The picture is mostly stable but not flat. Train loss, eval KL, and CE utilization are roughly invariant across the sweep --- the proxy losses see nearly the same training signal at $T = 1$k as at $T = 8$k under matched compression. Compact MCQ accuracy, however, drops modestly with absolute context length, from $\sim 84.5\%$ at $T = 1$k to $\sim 80.4\%$ at $T = 8$k. The per-domain breakdown shows the same pattern without a single collapsing domain: at $T=8$k, Code reaches $79.0\%$, while Financial, Gutenberg, and Legal remain at or above $75\%$. The takeaway is that compact-cache utility at fixed compression ratio is approximately invariant to absolute context length on the proxy losses, with a modest accuracy hit on the harder domains as $T$ grows. This supports the design decision to train at moderate $T$ and rely on iterative compaction (\S\ref{subsec:iterative}) for longer deployment contexts, rather than scaling $T$ in training proportionally to the longest deployment length we want to support.

\section{Latent Scaling and CE as a Performance Proxy}
\label{app:latent-scaling-ce-proxy}

\begin{figure}[h]
    \centering
    \includegraphics[width=\linewidth]{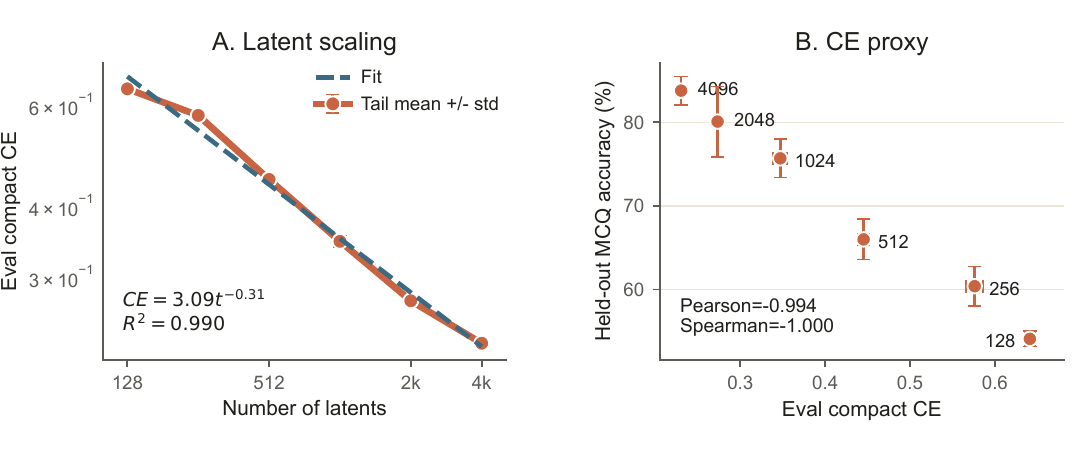}
    \caption{Latent-count scaling and compact CE as a downstream-performance proxy on the multidomain \(8192\)-token sweep. Left: compact CE falls smoothly as the number of latents increases, with a power-law-like fit over \(t \in \{128,256,512,1024,2048,4096\}\). Right: compact CE is strongly monotonic with held-out four-domain MCQ compact accuracy. Points and error bars are tail mean \(\pm\) std over the final \(50\%\) of evaluation checkpoints.}
    \label{fig:latent-scaling-ce-proxy}
\end{figure}

Figure~\ref{fig:latent-scaling-ce-proxy} tests two diagnostics on the cached multidomain latent sweep. First, compact CE improves smoothly with latent count: fitting \(\mathrm{CE}_{\mathrm{compact}} = a t^b\) gives \(b=-0.31\) with \(R^2=0.99\). This is not enough evidence for a headline scaling-law claim, but it shows that the trained compactors move along a stable quality--capacity curve rather than producing isolated operating points.

Second, compact CE is a strong proxy for held-out MCQ performance in this regime. Across the six latent counts, the four-domain aggregate MCQ accuracy increases monotonically as compact CE decreases, with Pearson correlation \(-0.994\) and Spearman correlation \(-1.000\). This supports using CE and KL as training-time model-selection signals for these sweeps, while keeping the conclusion scoped to a single-seed appendix diagnostic on one model and evaluation mixture.

\section{Compression-Ratio Cell-Level Comparisons}
\label{app:still-vs-am-delta}
 
The Pareto-style figure in \S\ref{subsec:design-space} compresses the
per-cell structure of the design-space sweep into a single frontier;
this appendix reports the cell-level views that support the qualitative
claims in the body. Figure~\ref{fig:compression-sweep} is the full
per-baseline compression-ratio sweep across context lengths,
Table~\ref{tab:exact-results-summary} gives the exact values behind the
Pareto plot, and Figure~\ref{fig:still-delta-panels} is the pair of
focused pairwise contrasts against the strongest baselines in each
non-Still cell of the design space: Attention Matching (per-context
synthesis) on the left and KV-Distill (amortized selection) on the
right.

\subsection{Targeted KVZip partial comparison}
\label{app:kvzip-partial}

KVZip \citep{kim2025kvzip} is the closest recent baseline to Still's
deployment regime among the missing neighbors because it scores cache
entries before the future user query is known. We therefore ported the
core cache-quality path into the same Qwen3-4B MCQ protocol used for
Table~\ref{tab:exact-results-summary}: for each document, KVZip's
context-reconstruction prompt assigns per-layer, per-KV-head token
importance scores, and the evaluator retains the top $K$ original cache
rows with direct values and no $\beta$ correction. This keeps the
retained-slot budget matched to the other compact-cache baselines, while
avoiding a confound from Still-specific value synthesis.

\begin{table}[h]
\centering
\scriptsize
\caption{Targeted KVZip partial comparison on the multidomain MCQ suite. KVZip
cells are compact-cache accuracy in percent on a $25$-document subset per
domain at the same cache budget $K$ used in the main design-space table; Still
and KV-Distill columns repeat the full-matrix means for reference.}
\label{tab:kvzip-partial}
\begin{tabular}{llrrrr}
\toprule
Ctx & Ratio & $K$ & Still & KV-Distill & KVZip \\
\midrule
8k  & 8x   & 1024 & 81.0 & 58.0 & 56.0 \\
8k  & 16x  & 512  & 78.0 & 53.0 & 43.0 \\
8k  & 32x  & 256  & 73.0 & 47.0 & 31.0 \\
8k  & 50x  & 164  & 62.0 & 50.0 & 25.0 \\
8k  & 100x & 82   & 55.0 & 42.0 & 22.0 \\
8k  & 200x & 41   & 51.0 & 45.0 & 21.0 \\
16k & 8x   & 2048 & 82.1 & 59.7 & 56.3 \\
16k & 16x  & 1024 & 66.6 & 54.0 & 34.8 \\
16k & 32x  & 512  & 59.3 & 50.7 & 26.3 \\
16k & 50x  & 328  & 53.6 & 46.1 & 23.8 \\
16k & 100x & 164  & 51.5 & 42.9 & 20.4 \\
16k & 200x & 82   & 50.5 & 43.0 & 22.7 \\
\bottomrule
\end{tabular}
\end{table}

The port above matches the cache-quality question
we need to answer for Still: if a query-agnostic selector is given the
same retained-cache budget, does it close the gap? On this subset it
does not; Still remains higher in every matched $8$k and $16$k cell, and
KVZip is only competitive with the selection baselines at low
compression. We do not run the full $8$k--$64$k matrix for three reasons.
First, this reference port omits KVZip's optimized sparse-serving kernels
and multi-query serving setup, so a full speed claim would not be a
faithful systems reproduction. Second, extending the reference implementation to every context length and every document would consume substantial additional evaluation time while adding little to the paper's main claim, so we report the scoped comparison explicitly and avoid claiming full-matrix dominance over KVZip.

\subsection{Prompted summarization baseline}
\label{app:summarization-baseline}

\begin{figure}[h]
    \centering
    \includegraphics[width=\linewidth]{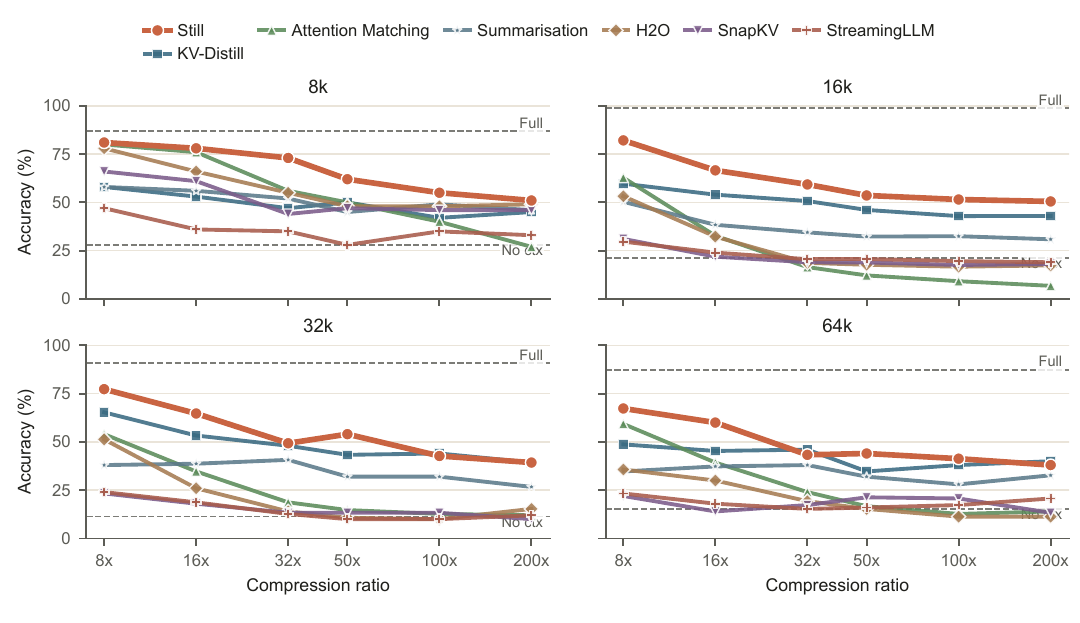}
    \caption{Per-cell compression-ratio sweep across context lengths. Compression ratio (log axis) vs.\ compact-cache MCQ accuracy at $\{8\text{k}, 16\text{k}, 32\text{k}, 64\text{k}\}$ contexts, all plotted methods, with full-context and no-context dashed references for each context length. This is the cell-level view behind the qualitative degradation claims in \S\ref{subsec:design-space}: heuristic selection (H2O, SnapKV) and StreamingLLM lose accuracy as the budget falls below the input's information density and as context length grows the eviction set wider; amortized selection (KV-Distill) sits above heuristic selection at every comparable cell but exhibits the same selection-bound floor; natural-language summarization is a slow lossy text bottleneck; Attention Matching exhibits a representational ceiling at extreme compression rather than a budget-bound failure; Still degrades gracefully across the matrix. For the learned methods (Still, KV-Distill), three thin lines per panel correspond to three training seeds; deterministic methods are drawn as a single line, with KVZip restricted to the $8$k/$16$k partial check in Appendix~\ref{app:kvzip-partial}. Per-cell SDs are reported in Table~\ref{tab:exact-results-summary}.}
    \label{fig:compression-sweep}
\end{figure}

The Summarization baseline in Figure~\ref{fig:pareto-frontier}, Figure~\ref{fig:compression-sweep}, and Table~\ref{tab:exact-results-summary} is a two-stage text bottleneck rather than a KV-cache method. For each document and compression budget \(K\), we first ask Qwen3-4B-Instruct-2507 to write a natural-language summary with at most \(K\) generated tokens. We then answer the same multiple-choice questions using the generated summary as the only context. The summary generation used temperature \(0.7\), top-\(p=0.8\), and top-\(k=20\); the answer pass used temperature \(0\) and a 16-token cap.

\begin{center}
\fcolorbox{black!18}{black!3}{%
\begin{minipage}{0.92\linewidth}
\footnotesize\ttfamily
Summarize the following text, providing all key points that might be necessary\par
to answer comprehension questions:\par\par
\{article\_text\}\par\par
Summary:
\end{minipage}}
\end{center}

For the answer pass, the model sees \texttt{Use the provided context to answer the multiple-choice question.}, followed by the generated summary, the question and answer options, and the instruction \texttt{Respond with ONLY the answer letter (A, B, C, or D). No explanation.}
 
The compression-ratio sweep makes the structural shape of each baseline's degradation directly visible. Heuristic selection methods (H2O, SnapKV) and StreamingLLM hold up at moderate compression on shorter contexts, where the retained subset is large enough to carry most of the answer-relevant information, but they degrade smoothly as the budget tightens and as context length grows --- the eviction set widens with $T$ at fixed compression ratio, and the selected subset has to span a longer document with the same number of slots. Amortized selection (KV-Distill) sits above heuristic selection at every comparable cell --- the learned scorer is meaningfully better than attention-mass thresholding --- but the curves have the same shape: at extreme compression on long context, no choice of original-token subset preserves enough information to recover full-context behavior. Prompted summarization gives the model an answerable text representation and often stays above heuristic selection, but it is slow and discards details before the answer pass. Attention Matching's curve has a different character. AM is competitive or better than Still at low compression on short contexts, but its degradation steepens sharply once compression and context length push the compact cache into the regime where per-context fitting cannot route enough distributed information through a small number of compact key rows. Still's curves are the flattest among the full-matrix compact-context methods, with the targeted KVZip check in Appendix~\ref{app:kvzip-partial} following the same selection-bound pattern on $8$k and $16$k.

\begin{table}[h]
\centering
\tiny
\setlength{\tabcolsep}{2.4pt}
\caption{Exact accuracy values behind Figure~\ref{fig:pareto-frontier}. Each row fixes the context length and compression ratio; $K$ is the resulting compact-cache budget. Entries are mean compact-cache MCQ accuracy with the standard deviation in parentheses, in percent. For learned methods (Still, KV-Distill) the SD is over $3$ training seeds; for deterministic methods (AM, AM-OMP, H2O, SnapKV, StreamingLLM) and the prompted summarisation baseline (Summ.) it is bootstrap over evaluation examples. KVZip entries are $25$-document partial runs at $8$k and $16$k and omit SDs; dashes denote settings not run. Full-context and no-context anchors are reported alongside Figure~\ref{fig:pareto-frontier} and the cell-level views in Figure~\ref{fig:compression-sweep} and are omitted from this table to make room for SDs. \textbf{Bold} indicates the best compact-cache mean per row.}
\label{tab:exact-results-summary}
\resizebox{\linewidth}{!}{%
\begin{tabular}{llrrrrrrrrrr}
\toprule
Ctx & Ratio & $K$ & Still & KV-Distill & AM & AM-OMP & Summ. & H2O & SnapKV & StreamingLLM & KVZip \\
\midrule
8k  & 8x   & 1024 & \textbf{81.0 (2.4)} & 58.0 (3.6) & 80.0 (1.4) & 68.0 (2.0) & 58.0 (1.9) & 78.0 (1.5) & 66.0 (1.4) & 47.0 (1.6) & 56.0 \\
8k  & 16x  & 512  & \textbf{78.0 (2.7)} & 53.0 (3.2) & 76.0 (1.6) & 69.0 (1.8) & 56.0 (2.1) & 66.0 (1.3) & 61.0 (1.7) & 36.0 (1.4) & 43.0 \\
8k  & 32x  & 256  & \textbf{73.0 (2.3)} & 47.0 (3.9) & 56.0 (1.3) & 60.0 (2.3) & 52.0 (1.7) & 55.0 (1.8) & 44.0 (1.3) & 35.0 (1.5) & 31.0 \\
8k  & 50x  & 164  & \textbf{62.0 (3.0)} & 50.0 (3.4) & 50.0 (1.7) & 48.0 (1.9) & 45.0 (2.3) & 48.0 (1.4) & 47.0 (1.5) & 28.0 (1.7) & 25.0 \\
8k  & 100x & 82   & \textbf{55.0 (2.6)} & 42.0 (4.1) & 40.0 (1.5) & 30.0 (2.1) & 49.0 (1.8) & 48.0 (1.6) & 46.0 (1.8) & 35.0 (1.3) & 22.0 \\
8k  & 200x & 41   & \textbf{51.0 (2.1)} & 45.0 (3.0) & 27.0 (1.2) & 22.0 (1.7) & 46.0 (2.2) & 49.0 (1.2) & 46.0 (1.4) & 33.0 (1.8) & 21.0 \\
16k & 8x   & 2048 & \textbf{82.1 (2.8)} & 59.7 (3.7) & 62.8 (1.8) & 60.0 (2.4) & 50.3 (1.6) & 53.2 (1.7) & 31.1 (1.6) & 29.7 (1.4) & 56.3 \\
16k & 16x  & 1024 & \textbf{66.6 (2.4)} & 54.0 (4.2) & 32.7 (1.4) & 47.8 (2.0) & 38.4 (2.0) & 32.3 (1.5) & 21.9 (1.2) & 24.0 (1.6) & 34.8 \\
16k & 32x  & 512  & \textbf{59.3 (3.1)} & 50.7 (3.3) & 16.5 (1.6) & 28.1 (1.8) & 34.5 (2.4) & 18.7 (1.3) & 19.0 (1.5) & 20.6 (1.2) & 26.3 \\
16k & 50x  & 328  & \textbf{53.6 (2.5)} & 46.1 (3.8) & 12.2 (1.3) & 21.0 (2.2) & 32.3 (1.8) & 17.7 (1.9) & 18.9 (1.7) & 20.7 (1.5) & 23.8 \\
16k & 100x & 164  & \textbf{51.5 (1.9)} & 42.9 (3.5) &  9.2 (1.9) & 13.5 (1.6) & 32.5 (2.1) & 16.7 (1.4) & 17.6 (1.3) & 19.6 (1.8) & 20.4 \\
16k & 200x & 82   & \textbf{50.5 (2.7)} & 43.0 (4.0) &  6.8 (1.5) &  9.4 (2.1) & 30.9 (1.9) & 17.4 (1.6) & 18.1 (1.8) & 19.0 (1.4) & 22.7 \\
32k & 8x   & 4096 & \textbf{77.3 (2.5)} & 65.3 (3.1) & 54.0 (1.4) & 45.0 (2.3) & 38.0 (2.3) & 51.3 (1.5) & 23.3 (1.4) & 24.0 (1.6) & -- \\
32k & 16x  & 2048 & \textbf{64.7 (3.0)} & 53.3 (3.6) & 34.7 (1.7) & 34.0 (1.9) & 38.7 (1.7) & 26.0 (1.7) & 18.0 (1.6) & 18.7 (1.3) & -- \\
32k & 32x  & 1024 & \textbf{49.3 (2.2)} & 48.0 (4.3) & 18.7 (1.2) & 28.0 (2.0) & 40.7 (2.0) & 14.0 (1.2) & 13.3 (1.5) & 12.7 (1.7) & -- \\
32k & 50x  & 655  & \textbf{54.0 (2.6)} & 43.3 (3.4) & 14.7 (1.5) & 22.0 (1.8) & 32.0 (2.2) & 10.7 (1.4) & 13.3 (1.3) & 10.0 (1.5) & -- \\
32k & 100x & 328  & 42.7 (2.9) & \textbf{44.0 (3.9)} & 12.7 (1.8) & 12.0 (2.2) & 32.0 (1.8) & 10.0 (1.8) & 13.3 (1.7) & 10.0 (1.4) & -- \\
32k & 200x & 164  & \textbf{39.3 (2.4)} & \textbf{39.3 (3.5)} & 12.0 (1.3) &  6.7 (2.0) & 26.7 (2.4) & 15.3 (1.5) & 10.0 (1.5) & 12.0 (1.6) & -- \\
64k & 8x   & 8192 & \textbf{67.3 (2.7)} & 48.7 (4.1) & 59.3 (1.6) & 50.0 (2.4) & 34.7 (1.6) & 35.7 (1.6) & 22.0 (1.4) & 23.3 (1.3) & -- \\
64k & 16x  & 4096 & \textbf{60.0 (2.3)} & 45.3 (3.2) & 39.3 (1.4) & 39.0 (1.7) & 37.3 (2.1) & 30.0 (1.3) & 14.0 (1.6) & 18.0 (1.8) & -- \\
64k & 32x  & 2048 & 43.3 (2.8) & \textbf{46.0 (3.7)} & 24.0 (1.7) & 31.0 (2.1) & 38.0 (1.9) & 19.3 (1.7) & 17.3 (1.2) & 15.3 (1.5) & -- \\
64k & 50x  & 1311 & \textbf{44.0 (2.5)} & 34.7 (3.4) & 16.7 (1.5) & 24.0 (1.9) & 32.0 (2.3) & 15.3 (1.4) & 21.3 (1.8) & 16.0 (1.7) & -- \\
64k & 100x & 655  & \textbf{41.3 (2.0)} & 38.0 (4.0) & 12.7 (1.3) & 14.0 (2.3) & 28.0 (1.7) & 11.3 (1.5) & 20.7 (1.3) & 17.3 (1.4) & -- \\
64k & 200x & 328  & 38.0 (2.6) & \textbf{40.0 (3.3)} & 14.0 (1.6) &  9.0 (1.8) & 32.7 (2.0) & 11.3 (1.6) & 13.3 (1.5) & 20.7 (1.5) & -- \\
\bottomrule
\end{tabular}%
}
\end{table}
 
\begin{figure}[h]
    \centering
    \includegraphics[width=\linewidth]{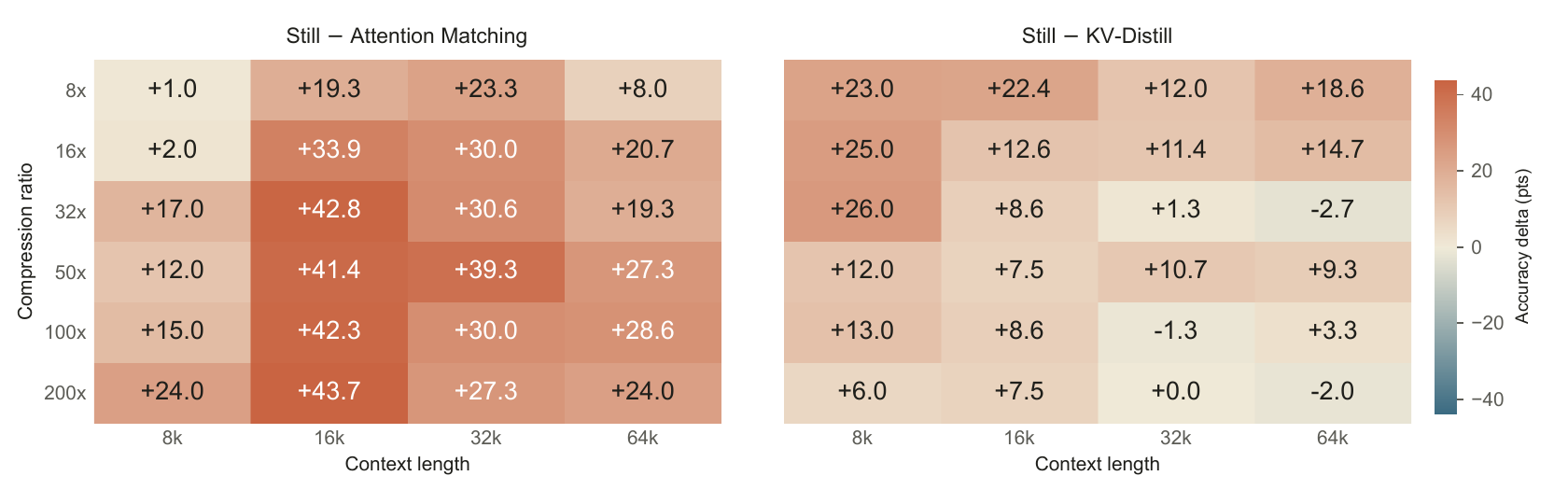}
    \caption{Still's accuracy advantage over Attention Matching (left) and KV-Distill (right) on the four-domain MCQ design-space matrix. Each cell reports the accuracy-point difference (Still minus baseline) at the same context length and compression ratio; both panels share the diverging color scale.}
    \label{fig:still-delta-panels}
\end{figure}

The Still-vs-AM panel (Figure~\ref{fig:still-delta-panels}, left) shows the gap small in the upper-left corner and growing with both compression ratio and context length, peaking in the $16$k--$32$k columns at extreme compression: at $8\times$ on $8$k the gap is roughly a single accuracy point in Still's favor, and by $32\times$ on $16$k it reaches $42.8$ points (Still $59.3\%$ vs AM $16.5\%$). The cleanest evidence that AM is hitting a representational ceiling rather than a budget-bound failure sits in the $16$k column at high compression, where AM falls to $6.8\%$ at $200\times$, $9.2\%$ at $100\times$, and $12.2\%$ at $50\times$ --- well below the $21.2\%$ no-context floor for that context --- while Still stays at $50.5\%$, $51.5\%$, and $53.6\%$ in the same cells. The Still-vs-KV-Distill panel has a different shape. At low compression the gap is large in Still's favor --- KV-Distill's selected subset cannot recover Still's compact-cache accuracy in those cells --- but at higher compression KV-Distill closes most of the gap, and in three cells crosses into a small positive lead, where its scorer's choice of retained tokens uses the available budget about as well as Still's compactor.
 
This is the cell-level evidence behind the §\ref{subsec:design-space} claim that AM's degradation is structural rather than budgetary. AM's per-context optimization is well-conditioned and computationally cheap at extreme compression --- the linear systems it solves shrink with $K$ --- so its falloff is not driven by running out of fitting budget. The mechanism is closer to a representational ceiling: the compact cache is geometrically the same as Still's, but AM's per-context fitting cannot route enough of a long context's distributed information through a small number of compact key rows for the frozen base model to recover the original answer distribution. Still uses the same compact-cache geometry but \emph{learns} to use it across a training distribution, and the resulting compactor synthesizes compact entries that AM's per-context objective leaves on the table.
 
The qualitative picture from Figure~\ref{fig:pareto-frontier} is that selection-based methods (heuristic and amortized alike) exhibit budget-bound rather than representation-bound degradation, with the breakdown point depending primarily on context length: at short context the selected subset can carry most of the answer-relevant information, and at long context no subset of the original tokens does. This is structurally different from AM's representational ceiling and from Still's graceful degradation. Table~\ref{tab:still-kvdistill-summary} summarizes the Still--KV-Distill cells that anchor the amortized synthesis versus amortized selection comparison.

\begin{table}[h]
\centering
\scriptsize
\caption{Still vs.\ KV-Distill result summary. Each entry reports the best compact-cache mean (SD) accuracy in percent over the swept cache budgets $K$ for that setting, with the selected $K$ in parentheses. SDs for the RULER matched-training, RULER transfer, and LongBench v2 rows are seed-level $\sigma$ over $3$ training seeds; the QuALITY-derived row reports single-seed bootstrap over evaluation examples. The RULER matched-training rows are the clean amortized synthesis versus amortized selection comparison for \S\ref{subsec:long-context}; the transfer rows are appendix diagnostics.}
\label{tab:still-kvdistill-summary}
\begin{tabular}{@{}p{0.30\linewidth}lccc@{}}
\toprule
Evaluation setting & Method & 32k & 64k & 128k \\
\midrule
RULER matched-training holdout & Still & 48.7 (2.7) ($K{=}2048$) & 39.9 (2.6) ($K{=}2048$) & 36.1 (2.5) ($K{=}2048$) \\
 & KV-Distill & 26.5 (3.4) ($K{=}2048$) & 21.9 (2.9) ($K{=}82$) & 20.3 (2.9) ($K{=}1024$) \\
\addlinespace
RULER transfer mixed-domain training & Still & 37.2 (2.6) ($K{=}2048$) & 32.9 (2.5) ($K{=}1024$) & 31.3 (2.5) ($K{=}1024$) \\
 & KV-Distill & 24.1 (3.2) ($K{=}2048$) & 21.8 (2.9) ($K{=}512$) & 22.0 (3.0) ($K{=}2048$) \\
\addlinespace
QuALITY-derived transfer & Still & 42.5 (1.8) ($K{=}2048$) & 42.0 (1.6) ($K{=}328$) & 33.8 (1.5) ($K{=}1024$) \\
 & KV-Distill & 40.0 (1.7) ($K{=}512$) & 44.2 (1.4) ($K{=}164$) & 36.5 (1.8) ($K{=}1024$) \\
\addlinespace
LongBench v2 context-dependent transfer & Still & 36.4 (2.6) ($K{=}1024$) & 58.3 (2.9) ($K{=}328$) & 16.7 (2.3) ($K{=}328/164$) \\
 & KV-Distill & 26.0 (3.1) ($K=1024$) & 25.0 (3.0) ($K{=}328$) & 16.7 (2.7) ($K=328/164$) \\
\bottomrule
\end{tabular}
\vspace{0.25em}
\begin{minipage}{0.92\linewidth}
\end{minipage}
\end{table}

\section{LongBench Generation Methodology}
\label{app:longbench-generation-method}

The LongBench generation audit in Table~\ref{tab:longbench-generation-pairwise} uses LongBench v1 \citep{bai2024longbench} GovReport and QMSum, evaluated with Qwen3-4B-Instruct-2507. We construct a fresh held-out 16k sample with seed $123$: for each task, we shuffle rows by stable LongBench ID and keep $250$ examples whose context is at least $8192$ Qwen tokens and shorter than $16384$ tokens (500 examples total). The same sampled manifest is used for Still and KV-Distill. GovReport uses the standard ``write a one-page summary'' prompt; QMSum uses the transcript plus query prompt. For compact-cache conditions, the prompt prefix and document context are prefetched and compacted, while the post-context task suffix remains on the uncompressed answer path.

Both methods use the same retained budget, $K=1024$, the same base model, and the identical generation-aligned continuation recipe. The Still and KV-Distill generation-aligned checkpoints are each short continuations of their respective 16k, $K=1024$ checkpoints on GovReport/QMSum summary rows: each retains the top-$200$ KL term against the uncompressed-cache teacher and adds answer-text cross-entropy on the reference summaries with weight $100$ for $50$ additional steps. Decoding is deterministic greedy generation with \texttt{max\_new\_tokens=192}, repetition penalty $1.2$, and no-repeat $4$-gram blocking.

We judge only the Still-vs-KV-Distill pair. For each example, the two outputs are anonymized and randomly assigned to output aliases before judging. The judge is \texttt{openai/gpt-5.5}; it sees the LongBench task input, the reference answer(s), and the anonymized candidate outputs, assigns each output a $0$--$100$ quality score for factual coverage, faithfulness, and task usefulness, and chooses the better output or a tie. No ties occur in the reported $500$ examples.

\section{Long-Context Evaluation Datasets}
\label{app:long-context-datasets}

All long-context evaluations use the Qwen3-4B-Instruct-2507 tokenizer and
record source revisions, tokenizer metadata, seeds, build commands, and row
counts in dataset manifests. Table~\ref{tab:long-context-datasets} summarizes
the four dataset families used for the 32k--128k evaluations. Long-MCQ is the
four-domain generated-MCQ suite used for the iterative-compaction headline; the
other three datasets stress complementary behaviors: concatenated real reading
comprehension, synthetic long-context retrieval and aggregation, and exact
needle retrieval.

\paragraph{Existing asset licenses.}
We use Qwen3 checkpoints under Apache-2.0 and Gemma-3 checkpoints under the
Gemma license. Evaluation and source datasets are credited to their creators:
QuALITY is distributed under CC BY 4.0; RULER is Apache-2.0; HELMET is MIT;
LongBench is MIT and LongBench v2 is Apache-2.0; PG19 is Apache-2.0, with its
underlying Project Gutenberg texts governed by the Project Gutenberg License;
and Stack-v1 is governed by The Stack terms of use and the original
permissive licenses attached to each source file. We use these assets for
research evaluation and derived dataset construction consistent with their
published terms; any released derived artifacts will include source
attribution, license notes, and links to the original assets.

\begin{table}[h]
\centering
\scriptsize
\caption{Long-context evaluation datasets. All contexts are measured with the
Qwen tokenizer. Generated-domain and QuALITY rows are multiple-choice; RULER
and Needle are open-answer.}
\label{tab:long-context-datasets}
\begin{tabular}{@{}p{0.14\linewidth}p{0.18\linewidth}p{0.15\linewidth}p{0.39\linewidth}@{}}
\toprule
Dataset & Source & Lengths & Construction \\
\midrule
Long-MCQ & Financial, Gutenberg, Legal, Code corpora & 32k, 64k, 128k &
Generated extractive MCQs from long natural documents using a Qwen teacher, with
no-context filtering and teacher verification. Each domain/length targets 50
documents; the Code split uses a Stack-v1 \citep{kocetkov2022stack} long-source regeneration to avoid
auto-generated, length-biased code rows. \\
\addlinespace
QuALITY-concat & \texttt{emozilla/quality} train split & 8k, 16k, 32k, 64k, 128k &
Concatenates eligible QuALITY articles to the target length, samples questions
only from fully included articles, preserves the native hard/easy flag, and
uses deterministic cycling when strict no-reuse supply is insufficient at long
lengths. \\
\addlinespace
RULER & Official NVIDIA/RULER generator & 32k, 64k, 128k &
Seven-task subset: \texttt{niah\_single\_2}, \texttt{niah\_multikey\_1},
\texttt{vt}, \texttt{cwe}, \texttt{fwe}, \texttt{qa\_1}, and
\texttt{qa\_2}. We use deterministic train/holdout splits for the
matched-training comparison and reserve transfer evaluations as diagnostics. \\
\addlinespace
Needle & PG19 \citep{rae2019compressive} prose and Stack-v1 code haystacks & 32k, 64k, 128k &
Deterministic needle insertion at depths \(\{0,10,25,50,75,90,100\}\%\) with
50 examples per cell. Prose uses a paragraph/sentence smart-snap variant to
avoid mid-sentence insertions; code keeps language-balanced Stack-v1 haystacks.
\\
\bottomrule
\end{tabular}
\end{table}

\section{Architecture Ablations}
\label{app:arch-ablations}

We probe the sensitivity of Still to small architectural choices under a
$\beta$-enabled diagnostic configuration (distinct from the $\beta=0$
setup of every reported Still result in the main paper; see
Appendix~\ref{app:architecture}): Qwen3-4B, $8192$-token context,
$1024$ latents ($8\times$ compression), $d_\ell=256$, two blocks, repeated
cross-attention, RoPE-fix, freely learned $\beta$, and answer-token KL training.
Each row in Table~\ref{tab:arch-ablation} is trained on the same four-domain
MCQ corpus with the same optimizer and schedule; values are tail-means over the
final $30\%$ of evaluation checkpoints. The \texttt{resmse} row is an objective
diagnostic under the same architecture, not an architecture ablation.

\begin{table}[h]
\centering
\caption{Canonical architecture ablations at $1024$ latents ($8\times$ compression) on the multidomain benchmark. Architecture rows modify one component of the canonical Still configuration; \texttt{resmse} changes only the objective. Down-arrows mark metrics where lower is better.}
\label{tab:arch-ablation}
\scriptsize
\begin{tabular}{llrrrr}
\toprule
Variant & Description & Eval KL $\downarrow$ & Eval CE $\downarrow$ & Util (CE) $\uparrow$ & MCQ acc $\uparrow$ \\
\midrule
\texttt{baseline} & Canonical Still & 0.302 & 0.416 & 0.836 & 0.708 \\
\texttt{nosa}     & Self-attention removed & 0.293 & 0.407 & 0.842 & 0.738 \\
\texttt{mlp}      & Self-attention replaced by MLP & \textbf{0.290} & \textbf{0.403} & \textbf{0.845} & 0.740 \\
\texttt{2head}    & Two cross-attention heads & 0.303 & 0.416 & 0.836 & 0.738 \\
\texttt{1blk}     & One perceiver block & 0.308 & 0.422 & 0.833 & \textbf{0.756} \\
\texttt{resmse}   & Residual-MSE objective diagnostic & 0.923 & 1.050 & 0.437 & \textbf{0.756} \\
\bottomrule
\end{tabular}
\end{table}

The KL-trained architecture rows are tightly clustered: aggregate MCQ accuracy
ranges from $0.708$ to $0.756$, and eval KL varies by only $\sim 0.02$ absolute.
Removing self-attention, replacing it with an MLP, doubling cross-attention
heads, and halving depth all leave Still in the same operating regime. We keep
the two-block self-attention configuration for consistency across the main
compression sweep, not because this $8\times$ ablation identifies it as
load-bearing.

The objective diagnostic separates extractive accuracy from distributional
matching. \texttt{resmse} ties the best MCQ accuracy ($0.756$), but its eval KL
and CE are roughly three times worse than the KL-trained rows, and CE
utilization falls to $0.437$. Direct cache-style objectives can preserve enough
content for MCQ answering while producing a compact cache whose output
distribution is much farther from the full-context teacher, so we keep KL as the
paper-facing objective.

\subsection{No inter-layer compactor channel}

The per-layer compactors share no information at forward time. We do not add a compactor residual stream between layers because the base model's residual stream already propagates content vertically: layer $\ell+1$'s cache is computed from a residual stream that incorporates layer $\ell$'s output, so an explicit inter-layer connection would largely duplicate that signal. The architectural ablations above support this --- removing self-attention, replacing it with an MLP, doubling cross-attention heads, or halving depth all leave headline metrics in the same operating regime at $8\times$ compression. Adding inter-compactor capacity would also serialize an otherwise embarrassingly parallel forward pass across $L$ layers, regressing on the speed dimension central to this work.

\subsection{Sweeping $\beta$}

\begin{figure}
    \centering
    \includegraphics[width=\linewidth]{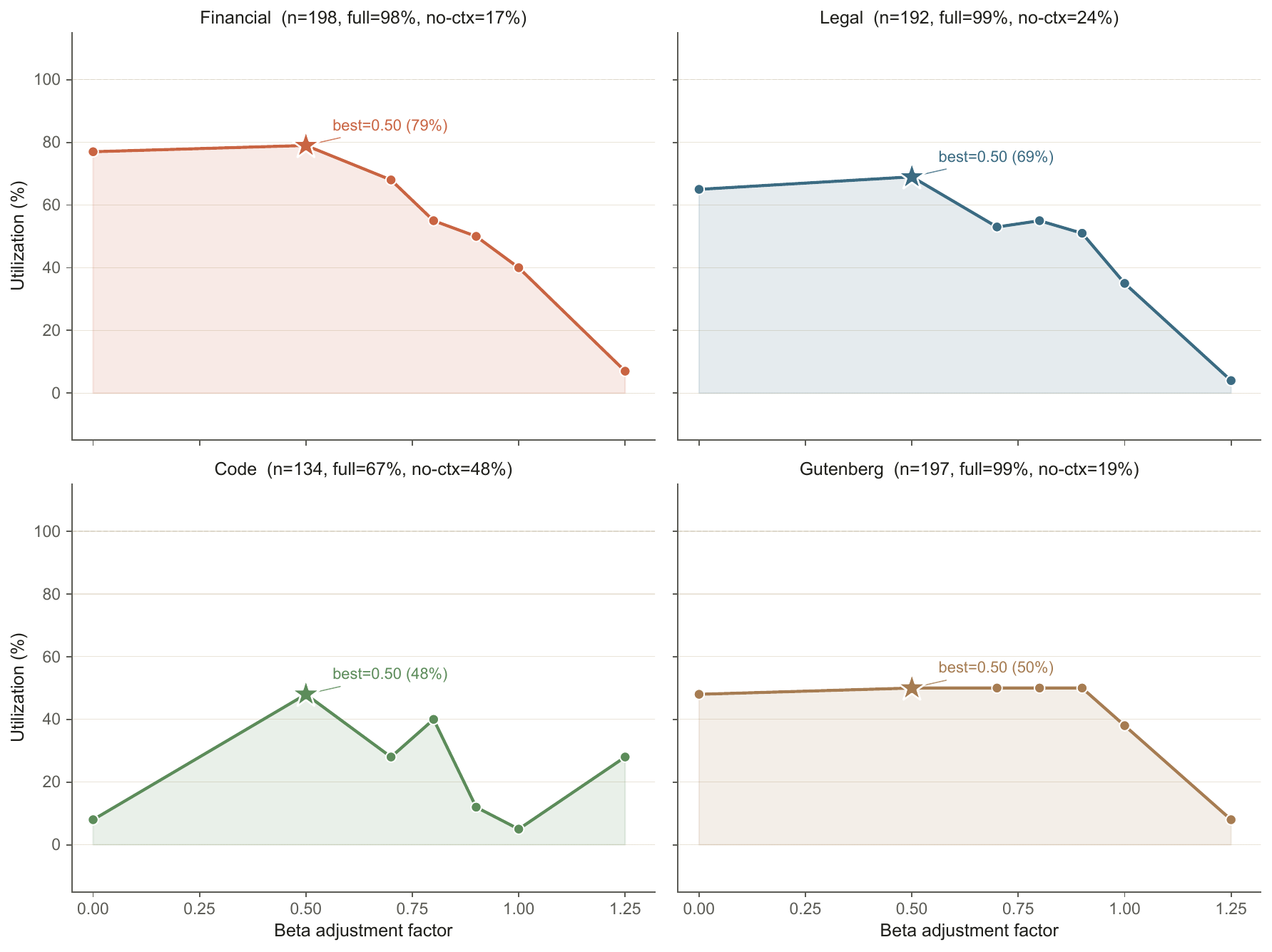}
    \caption{Sweeping the $\beta$ adjustment under iterative chunked compaction. The $1$k/$128$L checkpoint is applied iteratively to absorb an $8$k context in eight $1$k chunks, scaling the mass-matched correction $\beta_i = \hat{\beta}_i + \alpha \log(\text{chunks})$ by a factor $\alpha \in [0, 1.25]$. CE utilization (\%) is plotted per domain, with the per-domain best $\alpha$ marked by a star and the eval-set size, full-context, and no-context bounds in each subtitle. The optimum is consistently $\alpha \approx 0.5$ across domains: the $\log(\text{chunks})$ correction is in the right ballpark but overcorrects when applied at full strength, and $\alpha = 1.25$ collapses utilization in three of four domains.}
    \label{fig:beta_sweep_chunked}
\end{figure}

The mass-matched correction $\beta_i = \hat{\beta}_i + \log(T/t)$ that
Attention Matching uses, and that Still inherits as an optional prior
on its bias output (Appendix~\ref{app:architecture}), is calibrated
for single-pass compaction at a fixed compression ratio $T/t$. Under
iterative chunked compaction (\S\ref{sec:method-iterative}) the
effective ratio grows linearly in the number of passes, so the
natural question is whether the same logarithmic correction transfers.
We emphasize up front that this subsection is a diagnostic on a
$\beta$-re-enabled inference path: the headline iterative results in
\S\ref{subsec:iterative} all use $\beta=0$, consistent with every
other reported Still configuration in the paper
(Appendix~\ref{app:architecture}). Here we re-enable the optional bias
head on the same canonical $1$k/$128$-latent checkpoint to ask how
much calibration headroom $\beta$ would buy under recurrence if a
deployment supported it. We absorb an $8$k context in eight $1$k
chunks, scaling the correction as
$\beta_i = \hat{\beta}_i + \alpha \log(\text{chunks})$ for
$\alpha \in [0, 1.25]$, and track CE utilization per domain
(Figure~\ref{fig:beta_sweep_chunked}).

The optimum sits at $\alpha \approx 0.5$ across all four domains;
$\alpha = 1.25$ collapses utilization in three of the four. The
$\log(\text{chunks})$ form is the right shape but the magnitude
required at deployment is roughly half what the single-pass theory
prescribes: $\alpha = 0$ underweights the bias and the cache attends
too uniformly, $\alpha = 1$ overweights it and attention sharpens past
the optimum. The most plausible reading is that the compactor's own
$\hat{\beta}_i$ partially absorbs mass-matching across passes, leaving
the explicit additive correction redundant by half. The
$\alpha \approx 0.5$ optimum is what re-enabling $\beta$ would buy at
deployment; the reported iterative results in
\S\ref{subsec:iterative} sit at $\beta=0$ throughout, consistent with
every other Still number in the paper, so the gap to the per-domain
optima here is calibration headroom we leave on the table by choosing
not to take a deployment-incompatible dependency on the bias channel.
This $\alpha$ knob is precisely
the deployment-time calibration that \S\ref{sec:discussion} flags as
motivation for training Still directly under the iterative recurrence:
a compactor trained against its own compact-cache outputs would absorb
the correction into $\hat{\beta}_i$ and remove the only
deployment-tunable parameter in the system.

\subsection{Keys, values, and bias factorization}
\label{app:factorization-ablation}

This diagnostic enables the optional bias head and jointly produces compact keys $C_k$, compact values $C_v$, and per-token attention biases $\beta$. It asks which of the three learned outputs carries the quality gain. If selected keys plus synthesized values are nearly sufficient, then the main benefit of Still is value synthesis on top of ordinary token routing. If learned keys are also necessary, then the compactor is not merely filling selected memory slots with better content; it is learning the routing structure of the compact cache itself.

\paragraph{Setup.}
We train six Qwen3-4B compactors at three compression regimes,
$8192\rightarrow1024$ ($8\times$), $8192\rightarrow128$ ($64\times$),
and $8192\rightarrow41$ ($200\times$), holding the dataset, optimizer,
training budget, and evaluation protocol fixed. We report the matched
step-1500 checkpoint for each run. The beta-enabled baseline learns
$C_k$, $C_v$, and $\beta$. The remaining variants ablate one part of
this factorization: remove $\beta$ entirely; replace learned $\beta$
with the constant $\log(T/t)$; use learned keys but selected values; use
selected keys but learned values; or learn values only with selected
keys and uniform $\beta$. In the selected-key/value variants, selected
entries are the top-$k$ source positions by key RMS, so the comparison is
deliberately against a simple query-independent selection rule rather
than a tuned retrieval baseline.

%formatting of title with noctxt
\begin{figure}[h]
    \centering
    \includegraphics[width=\linewidth]{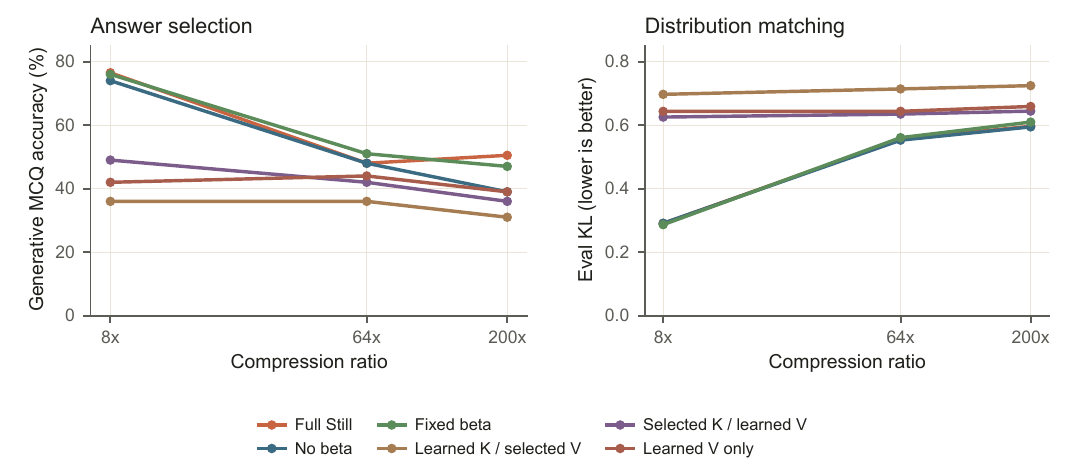}
    \caption{Keys/values/$\beta$ factorization ablation on Qwen3-4B across $8\times$, $64\times$, and $200\times$ compression, evaluated at the matched step-1500 checkpoint for each run. Left: compact-cache MCQ generation accuracy. Right: evaluation KL. Removing or fixing $\beta$ leaves the proxy close to $\beta$-enabled Still at low and medium compression, while replacing either learned keys or learned values with selected top-$k$ entries causes a large drop across all ratios.}
    \label{fig:factorization-ablation}
\end{figure}

\paragraph{Results.}
Figure~\ref{fig:factorization-ablation} shows that the learned bias is
not the main source of the gain at low or medium compression. At
$8\times$, $\beta$-enabled Still reaches $76.5\%$ compact MCQ accuracy and eval KL
$0.291$; no-$\beta$ and fixed-$\beta$ are essentially tied at
$74.0\%/0.291$ and $76.0\%/0.287$. At $64\times$, the same pattern
holds: $\beta$-enabled Still reaches $48.0\%$ and KL $0.556$, no-$\beta$ reaches
$48.0\%/0.553$, and fixed-$\beta$ reaches $51.0\%/0.561$. At the
extreme $200\times$ point, learned $\beta$ matters more for MCQ answer
generation: $\beta$-enabled Still reaches $50.5\%$ compact accuracy, compared with $39.0\%$
for no-$\beta$ and $47.0\%$ for fixed-$\beta$, while the KL proxy
remains close for all three ($0.596$, $0.595$, and $0.610$).

\paragraph{Both learned keys and learned values matter.}
The selection hybrids are the informative rows. Learned keys with
selected values is consistently weak, reaching only $36.0\%$ at
$8\times$, $36.0\%$ at $64\times$, and $31.0\%$ at $200\times$.
Selected keys with learned values recovers part of the gap
($49.0\%$, $42.0\%$, and $36.0\%$), and learned-values-only lands in the
same lower band ($42.0\%$, $44.0\%$, and $39.0\%$), but all remain well
below $\beta$-enabled Still at the matched ratios. These results support the
interpretation that values carry much of the recoverable content, but
learned keys are not dispensable: synthesized values help when routed
through selected keys, yet the best compact cache requires learning the
router and the content jointly. Still's advantage is not only that it
writes better values into a selected memory; it also learns compact keys
that organize how the frozen model will retrieve from that memory.

\section{RULER cell-level numbers}
\label{app:ruler-cells}

The matched-training and transfer RULER heatmaps in
Figure~\ref{fig:ruler-combined} use a shared color scale within each
panel. Per-cell 3-seed means and seed-level standard deviations are
reported in Table~\ref{tab:ruler-cells}; the matched-training rows
required additional per-cell training runs, while the transfer rows
re-evaluate the existing 3-seed Pareto checkpoints on RULER without
retraining. \emph{Within-noise} cells flagged in Figure~\ref{fig:ruler-combined}
are those whose Still--KV-Distill gap divided by its standard error
falls below $2$: matched training has two such cells (64k/$K{=}82$ and
128k/$K{=}82$); transfer has five
(32k/$K{=}82$, 64k/$K{=}82$, 128k/$K{=}328$, 128k/$K{=}164$, 128k/$K{=}82$).

\begin{table}[h]
\centering
\scriptsize
\setlength{\tabcolsep}{3pt}
\caption{Per-cell RULER accuracies behind Figure~\ref{fig:ruler-combined}, reported as 3-seed mean (seed-level $\sigma$) in percent. Top half: matched-training (compactor trained on RULER). Bottom half: zero-shot transfer from the mixed-domain MCQ corpus. \textbf{Bold} marks the best method per (length, $K$) cell.}
\label{tab:ruler-cells}
\begin{tabular}{llcccccc}
\toprule
Length & Method & $K{=}2048$ & $K{=}1024$ & $K{=}512$ & $K{=}328$ & $K{=}164$ & $K{=}82$ \\
\midrule
\multicolumn{8}{l}{\emph{Matched-training holdout}} \\
32k  & Still & \textbf{48.7 (2.7)} & \textbf{45.2 (2.7)} & \textbf{38.4 (2.6)} & \textbf{39.2 (2.6)} & \textbf{34.5 (2.5)} & \textbf{29.1 (2.5)} \\
     & KV-Distill & 26.5 (3.4) & 24.0 (3.3) & 22.1 (3.0) & 21.2 (3.0) & 19.0 (2.9) & 18.7 (2.9) \\
64k  & Still & \textbf{39.9 (2.6)} & \textbf{36.5 (2.6)} & \textbf{31.8 (2.5)} & \textbf{36.5 (2.6)} & \textbf{35.1 (2.5)} & \textbf{24.2 (2.4)} \\
     & KV-Distill & 21.0 (2.9) & 20.4 (2.9) & 21.1 (2.9) & 20.4 (2.9) & 20.2 (2.9) & 21.9 (2.9) \\
128k & Still & \textbf{36.1 (2.5)} & \textbf{32.9 (2.5)} & \textbf{27.4 (2.4)} & \textbf{30.2 (2.5)} & \textbf{31.3 (2.5)} & \textbf{19.3 (2.3)} \\
     & KV-Distill & 19.4 (2.9) & 20.3 (2.9) & 19.4 (2.9) & 17.9 (2.8) & 17.1 (2.8) & 17.6 (2.8) \\
\midrule
\multicolumn{8}{l}{\emph{Zero-shot transfer from mixed-domain MCQ training}} \\
32k  & Still & \textbf{37.2 (2.6)} & \textbf{34.0 (2.5)} & \textbf{29.6 (2.5)} & \textbf{33.7 (2.5)} & \textbf{33.3 (2.5)} & \textbf{22.1 (2.4)} \\
     & KV-Distill & 24.1 (3.2) & 22.7 (3.0) & 21.3 (2.9) & 22.0 (3.0) & 22.7 (3.0) & 18.8 (2.9) \\
64k  & Still & \textbf{31.8 (2.5)} & \textbf{32.9 (2.5)} & \textbf{27.5 (2.4)} & \textbf{31.7 (2.5)} & \textbf{27.5 (2.4)} & \textbf{20.6 (2.3)} \\
     & KV-Distill & 20.2 (2.9) & 21.0 (2.9) & 21.8 (2.9) & 19.3 (2.9) & 18.6 (2.8) & 19.4 (2.9) \\
128k & Still & \textbf{30.4 (2.5)} & \textbf{31.3 (2.5)} & \textbf{25.8 (2.4)} & \textbf{23.4 (2.4)} & 20.5 (2.3) & \textbf{18.9 (2.3)} \\
     & KV-Distill & 22.0 (3.0) & 19.5 (2.9) & 20.3 (2.9) & 19.8 (2.9) & \textbf{20.7 (2.9)} & 18.1 (2.9) \\
\bottomrule
\end{tabular}
\end{table}

\section{Iterative Compaction}
\label{app:iterative-compaction}

%We should give context on how each of these was trained, and how evaluated (ie what no-lookahead and concat-final mean)

This appendix reports the full iterative-compaction diagnostics. The main
paper focuses on Long-MCQ because it gives the cleanest long-context signal:
the same benchmark has stable full-context and no-context references at
32k/64k/128k, and the 16k- and 32k-trained compactors remain useful at 128k.
The appendix includes the broader aggregate sweep, QuALITY transfer,
RULER transfer, task-level RULER breakdowns, per-domain Long-MCQ results, and
an evaluation-mode ablation.

\begin{figure}[h]
    \centering
    \includegraphics[width=\linewidth]{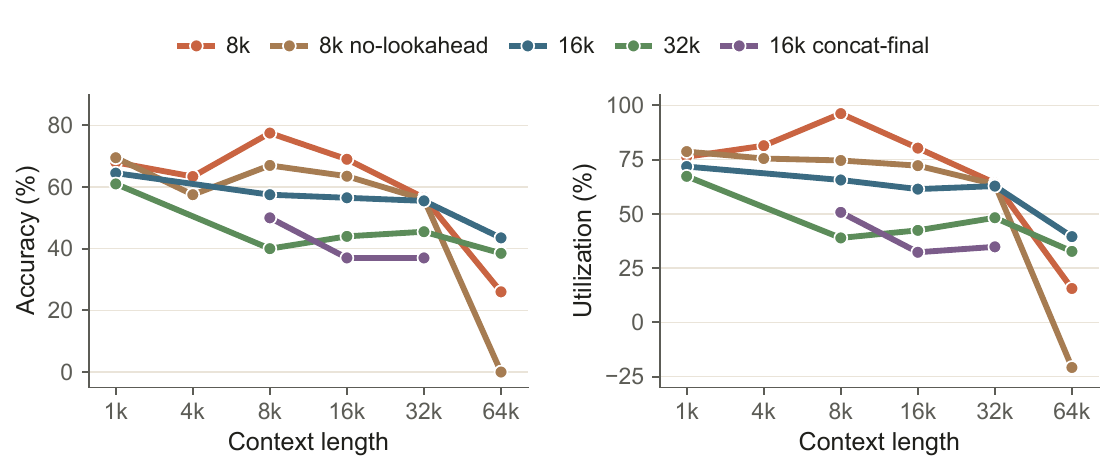}
    \caption{Aggregate iterative context sweep. Compact accuracy and
    utilization are shown across 1k--64k contexts for the main training
    families. The 8k and no-lookahead variants perform well near their
    training horizon but degrade at 64k; the 16k-trained compactor is more
    stable at long contexts, and the 32k-trained compactor trades lower
    short-context accuracy for better long-context behavior.}
    \label{fig:iter-aggregate-appendix}
\end{figure}

The aggregate sweep in Figure~\ref{fig:iter-aggregate-appendix} collects the
earlier context-sweep evaluations into one view. It is useful as a sanity
check but less clean than the Long-MCQ result because it mixes evaluation
surfaces and context ranges. The qualitative conclusion is consistent: models
trained only for short iterative horizons can look strong at 8k and 16k, but
the ordering changes at longer contexts.

\begin{figure}[h]
    \centering
    \includegraphics[width=\linewidth]{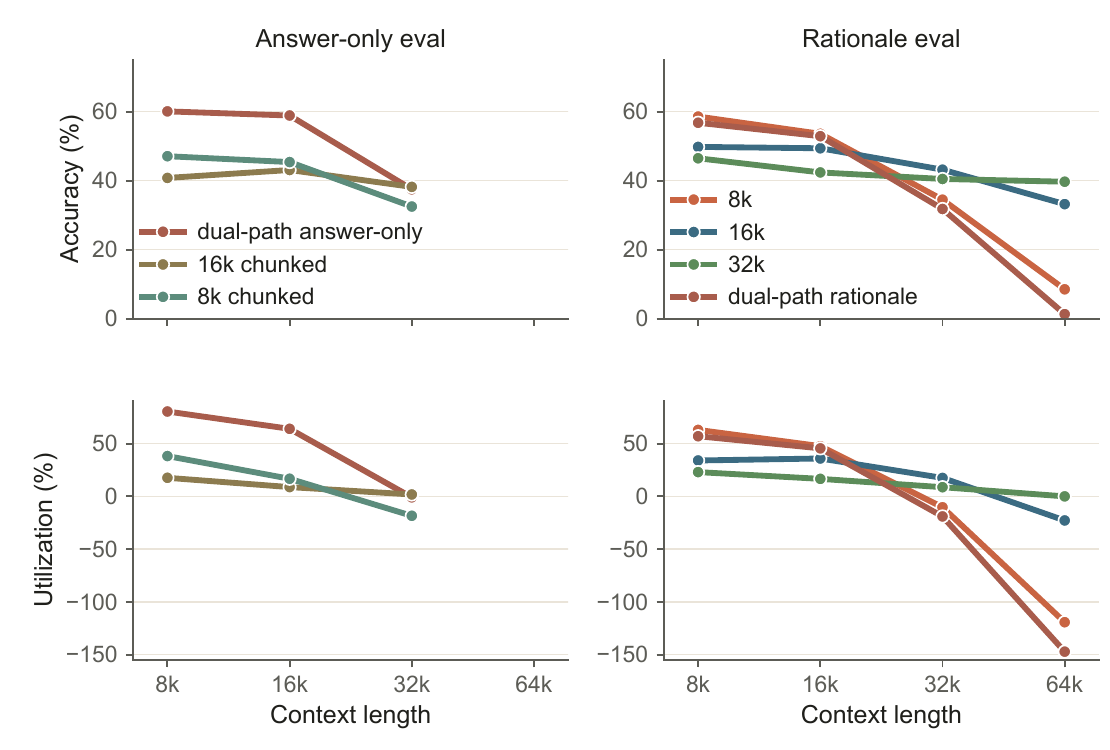}
    \caption{QuALITY-derived iterative sweep. We separate answer-only and
    rationale-style evaluations because they use different prompt surfaces.
    QuALITY performance is strong at 8k and 16k for the best answer-only and
    single-stage variants, but most iterative runs fall toward the no-context
    floor by 32k/64k. This makes QuALITY a useful stress test for repeated
    compaction, but not the cleanest headline result.}
    \label{fig:iter-quality-appendix}
\end{figure}

Figure~\ref{fig:iter-quality-appendix} shows that QuALITY is more brittle
under repeated compaction than Long-MCQ. The dual-path answer-only run is the
best short-context result, reaching about 60\% at 8k and 59\% at 16k, but it
drops to roughly the no-context floor at 32k. The 16k and 32k rationale runs
are more stable but lower overall. This supports two conclusions: prompt
surface matters for iterative compaction, and long-context QuALITY should be
used as a diagnostic rather than as the central online result.

\begin{figure}[h]
    \centering
    \includegraphics[width=\linewidth]{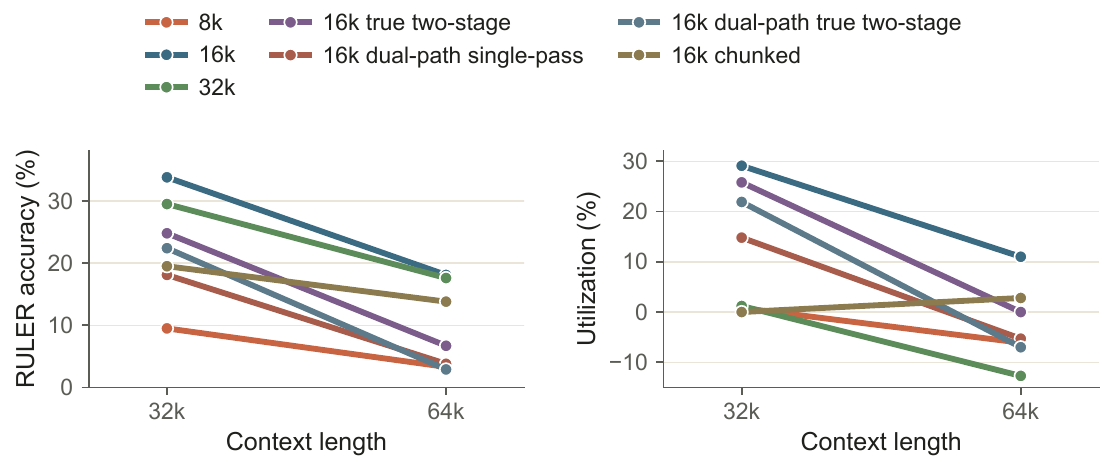}
    \caption{RULER aggregate iterative sweep. RULER is substantially harder
    than Long-MCQ under repeated compaction. The 16k-trained and 32k-trained
    lookahead compactors are the strongest aggregate variants, but absolute
    accuracy remains low and all methods degrade from 32k to 64k.}
    \label{fig:iter-ruler-appendix}
\end{figure}

RULER exposes the main weakness of the current iterative setup
(Figure~\ref{fig:iter-ruler-appendix}). The best aggregate 32k result is the
16k-trained lookahead compactor at 33.8\%, followed by the 32k-trained
lookahead compactor at 29.5\%. At 64k these fall to 18.1\% and 17.6\%.
Dual-path and true two-stage variants help on some individual tasks but do
not close the aggregate gap. We therefore report RULER as a diagnostic of
what remains difficult: repeated compaction can preserve broad MCQ evidence,
but exact synthetic retrieval and position-sensitive tasks are not solved by
the current training recipe.

\begin{figure}[h]
    \centering
    \includegraphics[width=\linewidth]{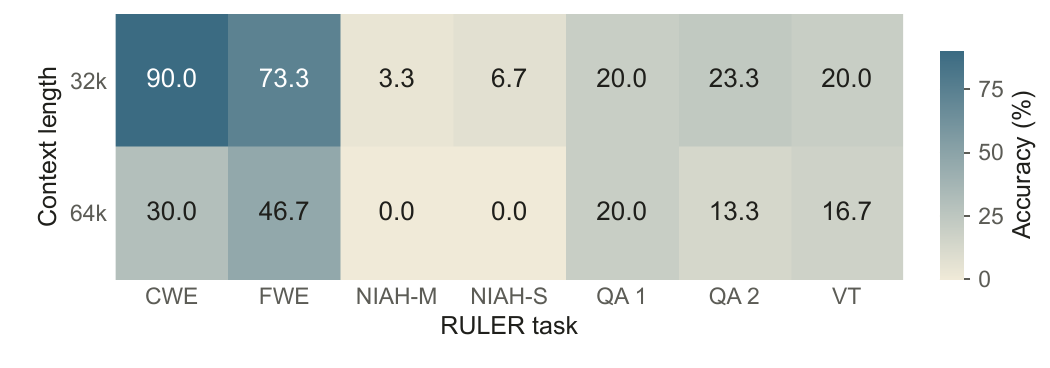}
    \caption{RULER task-level diagnostic for the strongest aggregate
    iterative checkpoint. Cells report compact-cache accuracy for the
    16k-trained lookahead compactor at 32k and 64k. The aggregate hides sharp
    task variation: CWE and FWE retain high accuracy at 32k, QA and VT retain
    partial signal, and needle-in-a-haystack variants mostly fail.}
    \label{fig:iter-ruler-task-heatmap}
\end{figure}

The task heatmap in Figure~\ref{fig:iter-ruler-task-heatmap} explains the
aggregate RULER behavior. The model can retain some structured evidence:
CWE reaches 90.0\% at 32k, FWE reaches 73.3\%, and QA/VT tasks remain above
zero. The needle tasks are different. Both single-needle and multi-key needle
variants are near zero at 32k and 64k. This suggests that the current
iterative compactor is not reliably preserving exact token strings through
many compression events, even when it retains enough semantic information for
less brittle tasks.

\begin{figure}[h]
    \centering
    \includegraphics[width=\linewidth]{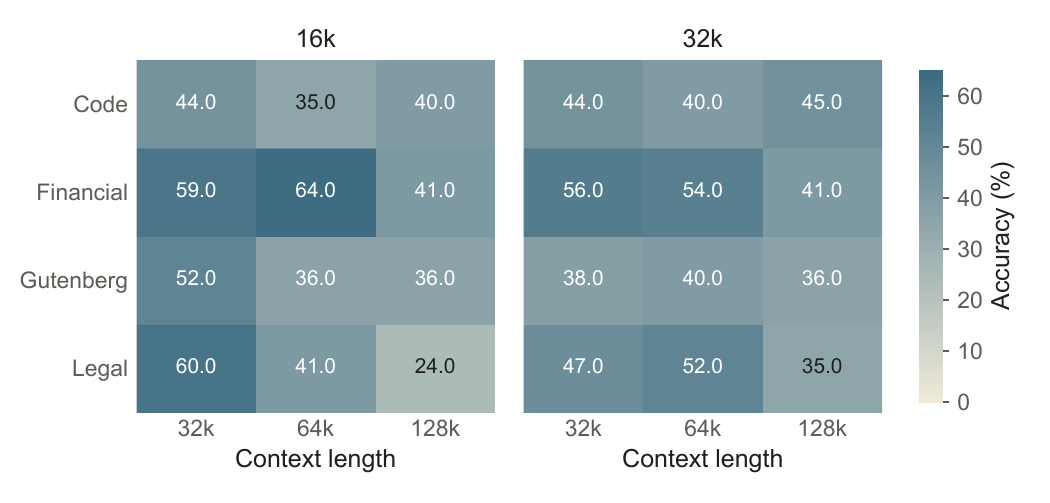}
    \caption{Long-MCQ domain breakdown for the 16k- and 32k-trained
    compactors. Each cell reports compact-cache accuracy by domain and context
    length. The 128k result is not a single-domain artifact: Code, Financial,
    Gutenberg, and Legal all retain nontrivial signal at 128k for at least one
    of the long-context-trained compactors.}
    \label{fig:iter-mudith-domain-heatmap}
\end{figure}

Figure~\ref{fig:iter-mudith-domain-heatmap} breaks the Long-MCQ result down
by domain. The 16k-trained compactor is strongest on Financial at 64k
(64.0\%) and remains nontrivial on Code, Financial, and Gutenberg at 128k.
The 32k-trained compactor is more balanced at 128k, reaching 45.0\% on Code,
41.0\% on Financial, 36.0\% on Gutenberg, and 35.0\% on Legal. The domain
view supports the main claim: the long-context result is distributed across
domains rather than driven by one easy subset.

\begin{figure}[h]
    \centering
    \includegraphics[width=\linewidth]{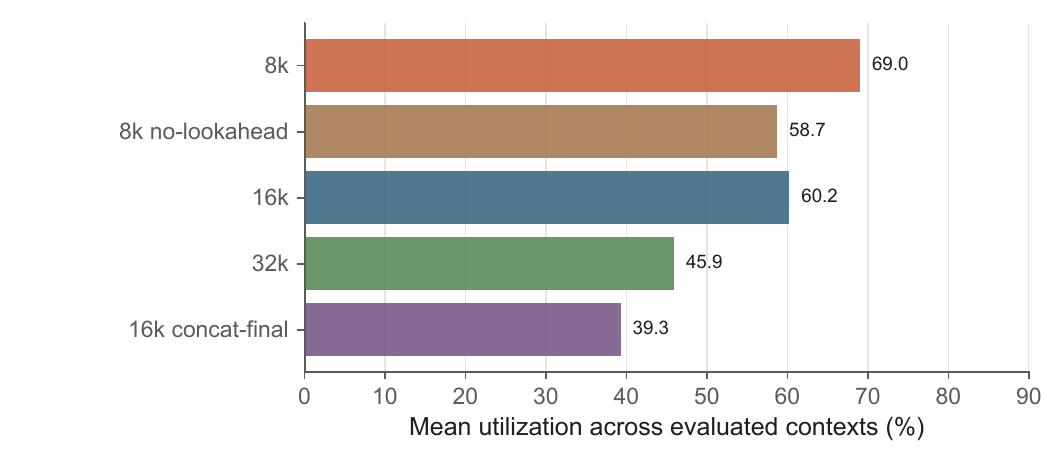}
    \caption{Evaluation-mode and training-family ablation. Bars report mean
    utilization across the evaluated contexts for each family. Single-stage
    lookahead is the most reliable setting in this sweep, no-lookahead is
    competitive near short horizons but less stable at 64k, and concat-final
    or true two-stage variants do not improve the aggregate result.}
    \label{fig:iter-eval-mode-ablation}
\end{figure}

The ablation in Figure~\ref{fig:iter-eval-mode-ablation} motivates the
curated main figure. Single-stage lookahead compactors trained at matched or
near-matched horizons are the most reliable family. No-lookahead retains
useful short-context performance but is less stable at long contexts.
Concat-final and true two-stage variants are useful engineering diagnostics,
but they do not improve the aggregate long-context result. This is why the
main figure reports the simple 8k/16k/32k lookahead ladder: it isolates the
effect of training horizon without mixing in secondary evaluation-mode
changes.

\section{Broader Impacts}
\label{app:broader-impacts}

Still is a systems method for reducing the retained KV cache of frozen language
models. Potential positive impacts include lower memory cost, lower serving
cost, and broader access to long-context inference on constrained hardware.
Potential negative impacts follow from making long-context language-model
deployment cheaper and more capable: the same efficiency gains could lower the
cost of harmful automation, surveillance over long records, or generation
systems that process sensitive documents. Still does not add new generative
capabilities to the base model, but responsible deployment should inherit the
base model's safety, privacy, and access-control safeguards and should treat
compacted caches as sensitive derived model state.

% \newpage
% \input{checklist.tex}
\end{document}